 \documentclass[final,3p,times,]{elsarticle}

\makeatletter
\def\ps@pprintTitle{%
  \let\@oddhead\@empty
  \let\@evenhead\@empty
  \def\@oddfoot{\reset@font\hfil\thepage\hfil} 
  \let\@evenfoot\@oddfoot}
\makeatother

\usepackage{amssymb}
\usepackage{amsmath}
\usepackage{amssymb}
\newcommand{\mb}[1]{\mathbf{#1}}
\newcommand{\bs}[1]{\boldsymbol{#1}}

\newcommand{\red}[1]{\textcolor{black}{#1}}

\usepackage{booktabs}
\usepackage{bbding}
\usepackage[table]{xcolor}
\definecolor{blue}{rgb}{0.21,0.49,0.74}
\usepackage[pagebackref,breaklinks,colorlinks,citecolor=blue]{hyperref}
\usepackage{wrapfig}
\usepackage{graphicx}
\usepackage{flushend}

\begin{document}

\begin{frontmatter}


\title{Learning Semantical Dynamics and SpatioTemporal Collaboration for Human Pose Estimation in Video} 

\author[1]{Runyang Feng}
\author[2]{Haoming Chen}

\affiliation[1]{organization={School of Artificial Intelligence, Jilin University},
            city={Changchun},
            postcode={130015}, 
            state={Jilin},
            country={China}}
            
\affiliation[2]{organization={School of Computer Science and Technology, East China Normal University},
            city={Shanghai},
            postcode={200062}, 
            country={China}}

\begin{abstract}
Temporal modeling and spatio-temporal collaboration are pivotal techniques for video-based human pose estimation.
 Most state-of-the-art methods adopt optical flow or temporal difference, learning local visual content correspondence across frames at the pixel level, to capture motion dynamics. However, such a paradigm essentially relies on localized pixel-to-pixel similarity, which neglects the \emph{semantical correlations} among frames and is vulnerable to image quality degradations (\emph{e.g.} occlusions or blur).  
 Moreover, existing approaches often combine motion and spatial (appearance) features via simple concatenation or summation, leading to practical challenges in fully leveraging these distinct modalities.
In this paper, we present a novel framework that learns multi-level semantical dynamics and dense spatio-temporal collaboration for multi-frame human pose estimation.
Specifically, we first design a Multi-Level Semantic Motion Encoder using a multi-masked context and pose reconstruction strategy. This strategy stimulates the model to explore multi-granularity spatiotemporal semantic relationships among frames by progressively masking the features of (patch) cubes and frames.
We further introduce a Spatial-Motion Mutual Learning module which densely propagates and consolidates context information from spatial and motion features to enhance the capability of the model.
Extensive experiments demonstrate that our approach sets new state-of-the-art results on three benchmark datasets, PoseTrack2017, PoseTrack2018, and PoseTrack21. 
\end{abstract}

\begin{keyword}
Human pose estimation \sep video-based human pose estimation \sep semantical motion modeling \sep deep learning 

\end{keyword}

\end{frontmatter}



\section{Introduction}
\label{sec:intro}

\red{Human pose estimation has long been a fundamental yet challenging task in the computer vision community. The goal is to localize human anatomical keypoints (\emph{e.g.}, wrist and ankle) for all persons in images or videos. {This task has received increasing attention in recent years~\cite{yuan2024multi, xu2023can} due to its successful applications in numerous scenarios including behavior understanding, augmented reality, and surveillance tracking~\cite{schmidtke2021unsupervised, yang2021learning}.}}

\begin{figure*}[t]
\begin{center}
\includegraphics[width=.94\linewidth]{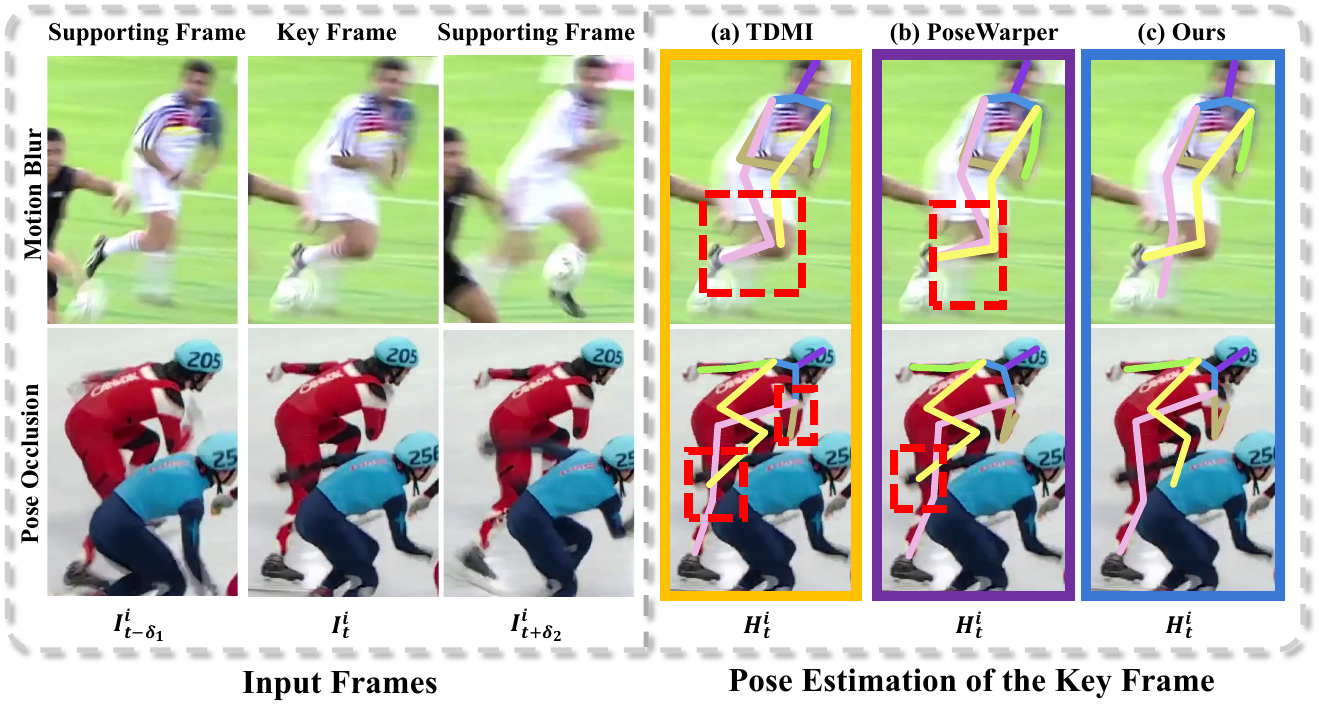}
\end{center}
\caption{
State-of-the-art methods like (a) TDMI \cite{feng2023mutual} and (b) PoseWarper \cite{bertasius2019learning} focus on modeling local pixel-wise dynamics based on feature similarities and they both show limitations during severe occlusion and blur. In contrast, our method (c) is more robust by fully exploiting the multi-level semantic motion contexts.
}
\label{fig:cover}
\end{figure*}

Extensive research has been conducted in recognizing human poses from \emph{stationary images}, ranging from early attempts~\cite{wang2013beyond} that leverage pictorial structure models or graphical models to recent methods~\cite{niu2023convpose, wang2020improving} that employ convolutional neural networks~\cite{he2016deep} or Vision Transformers~\cite{li2021tokenpose, dosovitskiy2020image}.
 Despite the superior performance in still images, applying such models to video sequences remains challenging. By nature, videos present a more intricate structure~\cite{wu2022motion} than images due to the presence of an additional temporal dimension. 
\red{Therefore, effectively grasping and utilizing temporal dynamics is desirable to facilitate pose estimation in videos.}

One line of work explicitly introduces motion representations such as optical flow~\cite{song2017thin, pfister2015flowing} or temporal difference~\cite{feng2023mutual} on top of a CNN backbone, to enhance the exploitation of video data. \red{TDMI~\cite{feng2023mutual} conducts multi-stage temporal difference modeling to extract per-pixel movements and aggregates spatial and temporal features via cascaded convolutions.}
The literature~\cite{song2017thin, pfister2015flowing} computes dense optical flow between every two frames, and leverages the flow-based motion fields to refine pose heatmaps temporally. Another line of work \cite{wang2020combining, liu2022temporal} considers implicit motion compensation. 
FAMI-Pose~\cite{liu2022temporal} employs deformable convolutions to align the features of multiple frames in a pixel-wise manner, and summates all aligned feature maps for pose estimation. 

\begin{figure*}[t]
\begin{center}
\includegraphics[width=0.94\linewidth]{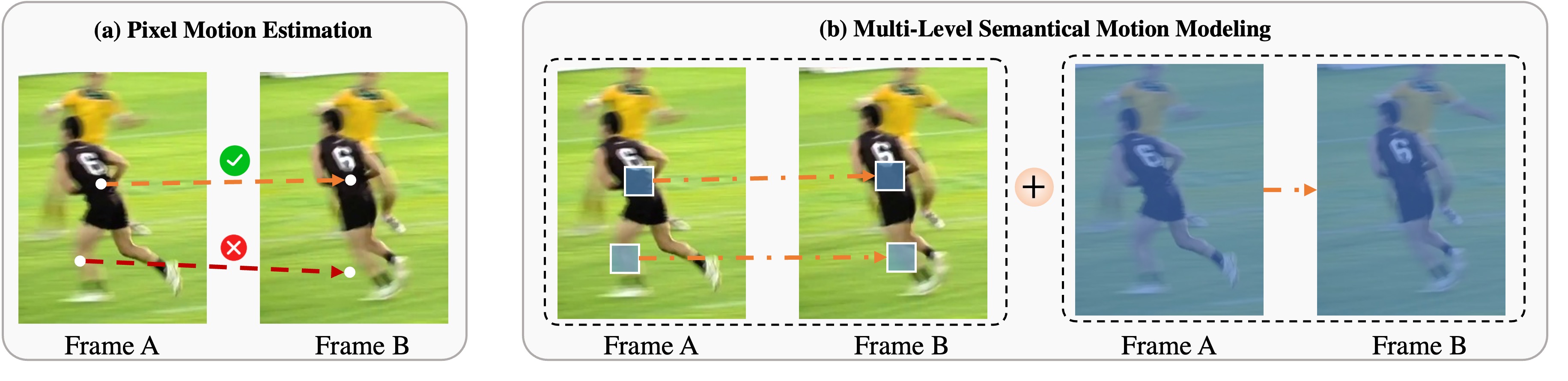}
\end{center}
\caption{\red{\textbf{Paradigm comparisons} of existing pixel-wise motion estimation and our proposed multi-level semantical motion modeling.}}
\label{fig:para}
\end{figure*}

By studying and experimenting with previous video-based human pose estimation (VHPE) methods~\cite{feng2023mutual, bertasius2019learning}, we empirically observe that they often suffer performance deterioration in challenging scenarios. As illustrated in Fig.~\ref{fig:cover}, prior well-established approaches like TDMI~\cite{feng2023mutual} inaccurately identify the ankle joints due to an improper understanding of human motions in severe blurred cases. These methods also encounter difficulties for occlusions, confusing the wrist and ankle joints that possess a similar appearance. We conjecture twofolds reasons that underline this phenomenon: \textbf{(i)} \red{Existing methods usually perform \emph{pixel-wise} motion estimation based on feature similarities to capture temporal dynamics. However, videos often involve cross-frame appearance inconsistencies due to the frequently occurred mutual occlusion or motion blur, which presents significant challenges for the pixel matching process~\cite{zhao2020towards} (as shown in Fig.~\ref{fig:para} (a)). Lacking in the capability of explicitly capturing  spatiotemporal semantic correlations among frames (\emph{e.g.}, human motion patterns), the performances of those methods are compromised especially for complex scenes.}
\textbf{(ii)} State-of-the-art approaches simply aggregate motion and appearance features through convolution or addition, which have difficulties in taking full advantage of these two complementary features and may obscure respective useful cues.

In this paper, we present a novel framework by jointly exploring multi-level \textbf{\underline{S}}emantic \textbf{\underline{D}}ynamics and dense Spatio\textbf{\underline{T}}emporal \textbf{\underline{C}}ollaboration  (\textbf{SDTC}) for VHPE. \textbf{(i)} \red{Masked signal modeling (\emph{i.e.}, masking certain input signals and attempting to recover these masked signals) allows for capturing relationships between signals and has been widely-used in natural language processing (NLP) and computer vision~\cite{xie2022simmim}.} Inspired by this, SDTC engages a Multi-Level Semantic Motion Encoder (MLSME) based on a multi-masked context and pose reconstruction strategy, which seeks to learn motion features from a hierarchical semantic affiliation perspective to overcome local pixel degradations. Specifically, MLSME progressively masks features of patch cubes and several frames within a sequence, and utilizes a patch- and frame-level motion encoder to extract corresponding dynamic representations, respectively. Then, the model learns to predict the feature contexts and pose heatmaps for masked locations (frames). \red{Through such a supplementary task of masked reconstruction during training, our MLSME can explore pose dynamics and delve into multi-level spatiotemporal semantic correlations among frames (Fig.~\ref{fig:para} (b)).}
\textbf{(ii)} SDTC further introduces a Spatial-Motion Mutual Learning (SMML) module to enhance the spatial and motion feature aggregation. It first refines cues within each modality, and subsequently densely exchanges context information between them based on cross-feature propagation. \red{Finally, SMML adaptively allocates pixel-wise attention weights to each modality to mutually aggregate them together.}

\red{We perform extensive evaluations on three large-scale benchmarks, including PoseTrack2017, PoseTrack2018, and PoseTrack21.} Experimental results show that SDTC delivers significant improvements over state-of-the-art methods. Our ablation studies further validate the efficacy of each proposed component and the design choice.

 Contributions of this work are summarized as: (1) We propose to tackle the task of video-based human pose estimation from the perspective of multi-level semantical motion modeling by leveraging multi-masked context and pose reconstruction. (2) We design a novel SMML which can effectively exploit spatial-motion aggregation to improve pose estimation performance.
 (3) \red{We demonstrate that SDTC sets new state-of-the-art results on three popular benchmark datasets, PoseTrack2017, PoseTrack2018, and PoseTrack21.}

 \section{Related Work}
\label{sec:related}

\subsection{Human Pose Estimation in Images}
 	\red{With the recent advancements in neural architectures, the deep learning models (\emph{e.g.}, CNNs~\cite{he2016deep} or Transformers~\cite{dosovitskiy2020image, vaswani2017attention}) have dominated various computer vision tasks such as salient object detection~\cite{wang2022hybrid}, action recognition~\cite{cheng2022tallformer}, and human pose estimation~\cite{xu2023can, zhou2024human}.} 	
 	\red{The deep learning-based pose estimation methods can be broadly divided into two streams: bottom-up~\cite{du2022hierarchical} and top-down~\cite{fang2017rmpe}.} \textbf{(i)} \emph{Bottom-up approaches} first detect all individual body joints and then group them to form the entire human pose. OpenPose~\cite{Cao_2017_CVPR} proposes a dual-branch framework that employs cascaded convolutions to localize body joints and affinity fields to encode part-to-part associations. PifPaf~\cite{kreiss2019pifpaf} leverages a Part Intensity Field to detect human body parts and designs a Part Association Field to associate body parts with each other. \textbf{(ii)} \emph{Top-down approaches} first detect bounding boxes for all persons and then estimate the human pose within each bounding box region. HRNet~\cite{wang2020deep} introduces a high-resolution network that maintains high-resolution feature maps in all network stages. TokenPose~\cite{li2021tokenpose} proposes token representations to explicitly learn the anatomical constraints between every two joints. ViTPose~\cite{xu2023vitpose++} employs plain vision transformers to extract strong representations for pose estimation, demonstrating superior performance in multiple benchmarks.  SUNNet~\cite{xu2021sunnet} employs human parsing information to improve the performance of pose estimation. MSPose~\cite{yuan2024multi} leverages multiple supervision to explore data-limited human pose estimation.

 \subsection{Human Pose Estimation in Videos}
  Existing image-based models struggle to handle video inputs effectively as they cannot utilize temporal information across frames~\cite{liu2021deep}. To tackle this problem, several studies propose to introduce temporal representations on top of a CNN backbone. TDMI~\cite{feng2023mutual} adopts temporal feature differences to model pixel motions and employs convolutions to aggregate motion and appearance features. Flow-based methods~\cite{song2017thin, pfister2015flowing} compute dense optical flow among frames and utilize such flow-based clues to refine the heatmap estimation. \red{DCPose~\cite{liu2021deep} and PoseWarper~\cite{bertasius2019learning} compute pixel-wise motion offsets between different frames and leverage motion fields to guide accurate pose resampling.} 
   Another line of literature \cite{wang2020combining, liu2022temporal} introduces implicit motion compensation. FAMI-Pose~\cite{liu2022temporal} proposes a framework which first aligns the features of each supporting frame to the keyframe at the pixel level, and then summates the overall feature maps to estimate pose heatmaps. 
  
  \red{ As the above methods strongly rely on pixel-level dynamics and neglect semantic motion patterns, they are particularly vulnerable to image quality degradations such as occlusion or blur. Furthermore, these methods crudely fuse motion and spatial features, which cannot fully leverage these two complementary modalities. In this paper, we aim to design a novel temporal modeling paradigm to learn multi-level semantical motion dynamics that are more robust to pixel degradations. On the other hand, inspired by previous works that focus on fully integrating multi-source information (\emph{e.g.}, SDNet~\cite{liu2023transcending} fuses scene clues and object information, SRAL~\cite{liu2023distilling} combines knowledge of super-resolution and salient detection), we propose a dense spatio-temporal collaboration strategy to take full advantage of motion and spatial features for VHPE.   
}

 \begin{figure*}
\begin{center}
\includegraphics[width=.97\linewidth]{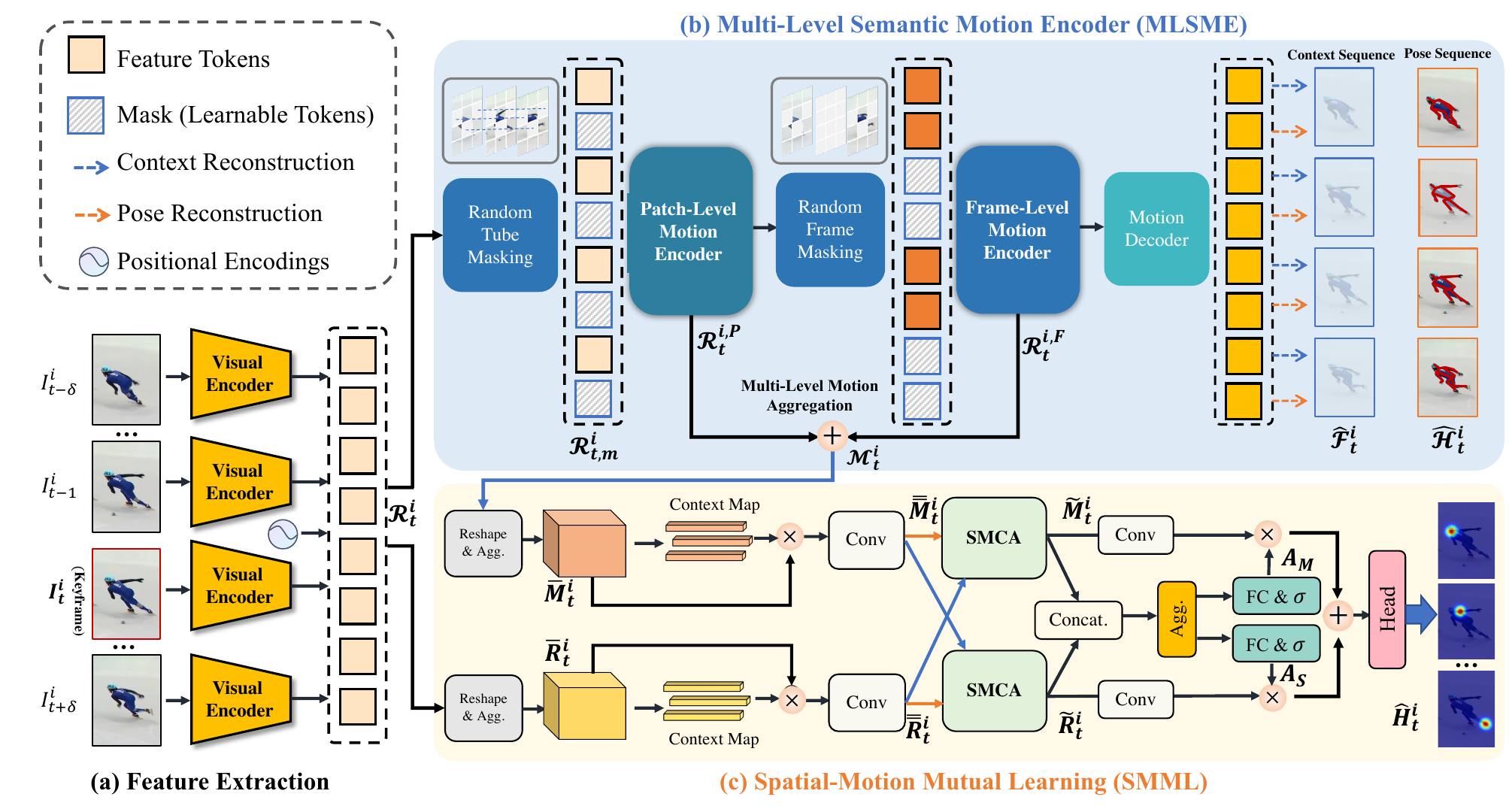}
\end{center}
\caption{\textbf{Overall pipeline} of our proposed framework. The goal is to estimate the human pose in the keyframe. Given the input sequence, we first extract their spatial features using a visual encoder. The resulting feature tokens are then processed via two modules (b) MLSME and (c) SMML for motion feature extraction and spatial-motion feature aggregation. Finally, a detection head is employed to produce the final pose estimation $\hat{\mb{H}}_t^{i}$.}
\label{fig:pipeline}
\end{figure*}

\section{Our Approach}
\label{sec:approach}

\noindent
\textbf{Preliminaries.}\quad
We take a  top-down approach in which a human detector~\cite{liu2021deep} is first used to extract the bounding boxes for all persons from a video frame $I_t$. 
Then, we enlarge each of the bounding boxes by $25\%$ to crop the individual $i$ across a frame sequence $\boldsymbol{\mathcal{I}_t^i} = \left \langle I_{t-\delta}^i,...,I_t^i,...,I_{t+\delta}^i \right \rangle$ with $\delta$ being a temporal span. Our goal is to fully exploit the temporal dynamics and spatiotemporal collaboration within the input sequence $\boldsymbol{\mathcal{I}_t^i}$ to estimate the human pose in the keyframe $I_t^i$.

\noindent
\textbf{Method overview.}\quad
The overall pipeline of the proposed SDTC is illustrated in Fig.~\ref{fig:pipeline}. There are two key components: Multi-Level Semantic Motion Encoder (MLSME) and Spatial-Motion Mutual Learning (SMML). Specifically, we first extract the visual features for each frame within $\boldsymbol{\mathcal{I}_t^i}$. Then, these features are successively processed by MLSME and SMML for multi-level semantic motion modeling and dense spatiotemporal collaboration. Finally, a detection head is used to obtain the final result $\hat{\mb{H}}_t^{i}$. In the following, we introduce the proposed MLSME and SMML in detail.

\subsection{Multi-Level Semantic Motion Encoder}
\label{sec:MLSME}

\red{We observe that optical flow or temporal difference has been widely used for temporal (motion) modeling in VHPE. However, this paradigm tends to rely on feature similarities to capture localized pixel dynamics, which is inevitably vulnerable to image quality degradations such as occlusion and blur.} Instead, understanding the semantic motion patterns of a sequence is promising to remedy this issue. On the other hand, masked signal modeling has demonstrated significant potential in extracting relations among signals.
Motivated by these analyses, we introduce the Multi-Level Semantic Motion Encoder (MLSME) to learn hierarchical semantical dynamics through a multi-masked context and pose reconstruction strategy.
\red{Our MLSME progressively masks feature tokens of patch cubes and frames, and recovers feature contexts and pose heatmaps for masked locations/frames based on the learned multi-granularity inter-frame correlations during training.}
 It contains three key steps, feature embedding extraction, patch-level motion encoding, and frame-level motion encoding.

\noindent
\textbf{Feature embedding extraction.}\quad 
Given the input sequence $\boldsymbol{\mathcal{I}_t^i} = \left \langle I_{t-\delta}^i,...,I_t^i,...,I_{t+\delta}^i \right \rangle$, we employ Vision Transformers~\cite{xu2023vitpose++} pretrained on COCO as the backbone to extract 1D embedding feature tokens for each frame. Considering that image sequence modeling is sensitive to both space and time locations, two types of positional encodings including a sine-cosine spatial embedding~\cite{vaswani2017attention} and a learnable temporal embedding are added to each token to yield the feature sequence $\boldsymbol{\mathcal{R}}_t^i = \left \langle R_{t-\delta}^i,...,R_t^i\in \mathbb{R}^{L\times C},...,R_{t+\delta}^i \right \rangle$, where $L$ and $C$ denote the number of tokens and channels, respectively.
\setcounter{figure}{3}
  \begin{figure*}
\begin{center}
\includegraphics[width=1.\linewidth]{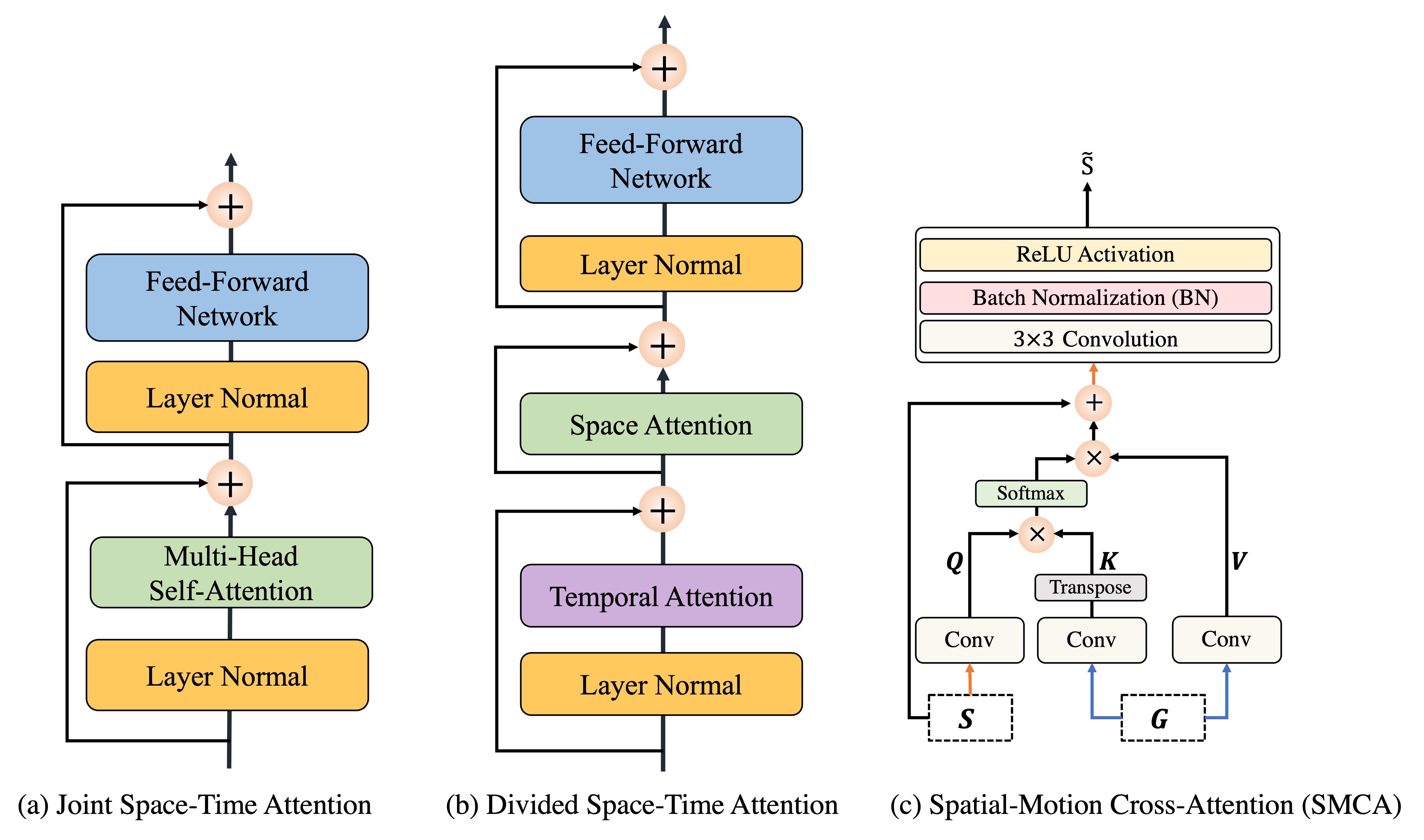}
\end{center}
\caption{\red{\textbf{Detailed structures} of the sub-components, including (a) joint space-time attention, (b) divided space-time attention, and (c) Spatial-Motion Cross-Attention (SMCA).}}
\label{fig:atten}
\end{figure*}

\begin{figure}[t]
\begin{center}
\includegraphics[width=0.9\linewidth]{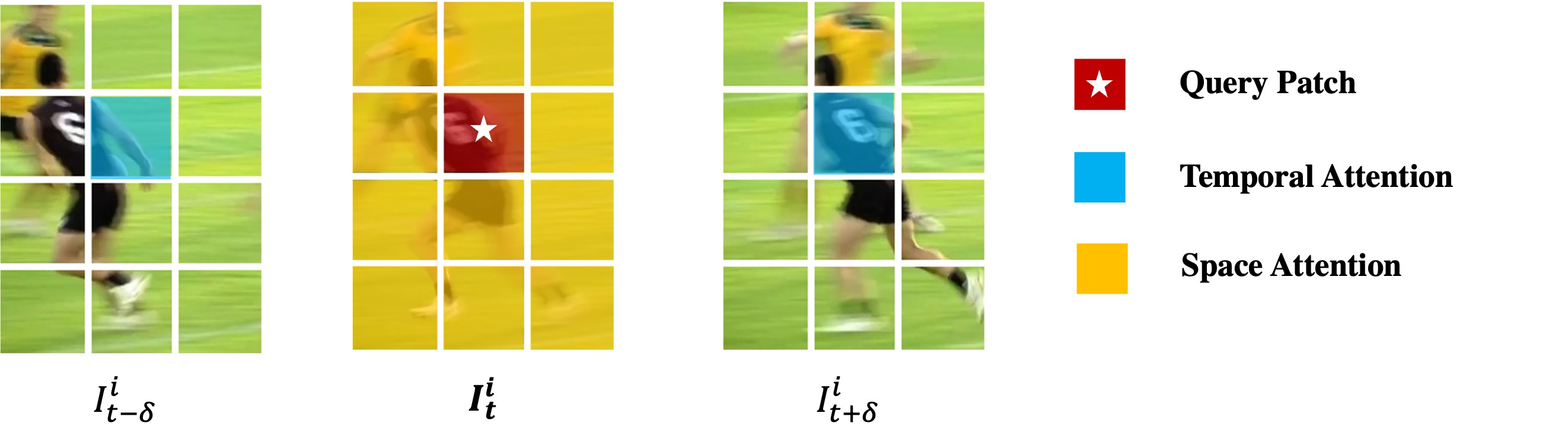}
\end{center}
\caption{\red{Visualization of \textbf{temporal attention and space attention schemes}. The patch in red indicates an arbitrary query patch within the input sequence, while blue and yellow patches represent the corresponding feature activations for temporal attention and space attention, respectively.}}
\label{fig:statten}
\end{figure}

 \noindent
 \textbf{Patch-level motion encoding.}\quad 
As illustrated in Fig.~\ref{fig:pipeline}, we design the patch-level motion encoder which explores inter-frame semantic relationships at the patch level to obtain motion features $\boldsymbol{\mathcal{R}}_{t}^{i,P}$. Specifically, given $\boldsymbol{\mathcal{R}}_t^i$, we first perform random temporal tube masking which enforces a mask to expand along the whole temporal axis (\emph{i.e.}, diverse frames sharing the same masking locations), producing the tensor $\boldsymbol{\mathcal{R}}_{t,m}^i$. Note that the masked patches are replaced with learnable token embeddings following the convention of patch (token) masking~\cite{xie2022simmim}. The above operation is expressed as:
  \begin{equation}
  	\begin{aligned}  
  	\boldsymbol{\mathcal{R}}_{t,m}^i = M_c * \boldsymbol{\mathcal{R}}_t^i + \hat{M}_c * L_c,
  	\end{aligned}
  \end{equation}
where $M_c$ denotes the random tube mask, $\hat{M}_c$ is the corresponding complementary mask, and $L_c$ indicates the learnable token embedding. Then, the feature tokens of each frame within $\boldsymbol{\mathcal{R}}_{t,m}^i$ are concatenated in the length dimension and fed into the  patch-level motion encoder which is composed of mixed joint space-time attention and divided space-time attention~\cite{bertasius2021space} layers. \red{The joint space-time attention can be implemented with the \emph{vanilla} multi-head self-attention~\cite{vaswani2017attention}, as shown in Fig.~\ref{fig:atten} (a), which enables all tokens to interact with each other and outputs the feature tensor $\boldsymbol{\mathcal{R}}_{t,m}^{i,J}$:}
 \begin{equation}
  	\begin{aligned}  
  	\boldsymbol{\mathcal{R}}_{t,m}^{i,J} &= \mathbf{MHSA}\left(\mathbf{Concat}_L\left(\boldsymbol{\mathcal{R}}_{t,m}^i\right)\right).
  		  	\end{aligned}
  \end{equation}
\red{To reduce the computational overhead, we further construct divided space-time attention layers (Fig.~\ref{fig:atten} (b)) to efficiently extract the motion representation $\boldsymbol{\mathcal{R}}_{t}^{i,P}$. In particular, we first perform \emph{temporal attention} over $\boldsymbol{\mathcal{R}}_{t,m}^{i,J}$ by computing the feature activations between each token and all tokens at the same spatial location in other frames, as depicted by the blue patches in Fig.~\ref{fig:statten}. The resulting feature encoding is then fed back for \emph{spatial attention} which captures feature interactions between each token and other tokens within the same frame (yellow patches in Fig.~\ref{fig:statten}), followed by a Feed Forward Network (FFN) to produce $\boldsymbol{\mathcal{R}}_{t}^{i,P}$.} This computation is formulated as:
  \begin{equation}
  	\begin{aligned}  
  		\boldsymbol{\mathcal{R}}_{t}^{i,P} &= \underbrace{{\mathbf{FFN}\left( \mathbf{SA}^+\left(\mathbf{TA}^+(\boldsymbol{\mathcal{R}}_{t,m}^{i,J})\right)\right)}}_{\text{divided space-time attention}},
  	\end{aligned}
  \end{equation}
 where $\mathbf{TA}^+$ and $\mathbf{SA}^+$ denote the temporal attention and the spatial attention with residual connections. \red{The temporal and spatial attentions can be implemented by modifying the computation dimensions of multi-head self attention to time and space, respectively.}

 \noindent
 \textbf{Frame-level motion encoding.}\quad After obtaining the patch-level motion features $\boldsymbol{\mathcal{R}}_{t}^{i,P}$, we further excavate frame-level spatiotemporal correlations to yield the corresponding motion representations $\boldsymbol{\mathcal{R}}_{t}^{i,F}$. Specifically, we first perform random frame masking over $\boldsymbol{\mathcal{R}}_{t}^{i,P}$, and similarly replace the masked tokens with learnable embedding vectors:
\begin{equation}
  	\begin{aligned}  
  	\boldsymbol{\mathcal{R}}_{t,fm}^{i,P} = M_f * \boldsymbol{\mathcal{R}}_{t}^{i,P} + \hat{M}_f * L_f,
  	\end{aligned}
  \end{equation}
where $ M_f$ is the random frame mask and $L_f$ denotes the learnable embedding. Subsequently, the feature $\boldsymbol{\mathcal{R}}_{t,fm}^{i,P}$ is passed into the frame-level motion encoder which outputs the motion feature $\boldsymbol{\mathcal{R}}_{t}^{i,F}$. The architecture of the frame-level motion encoder remains identical to the patch-level motion encoder, which can account for both sufficient token interactions and efficient computations. 

Given the hybrid-masked feature $\boldsymbol{\mathcal{R}}_{t}^{i,F}$, we perform masked reconstruction to enforce the patch- and frame-level motion encoders to discover more inter-frame semantical correlations. We feed $\boldsymbol{\mathcal{R}}_{t}^{i,F}$ into a motion decoder consisting of two multi-head self-attention layers, followed by separate MLP heads to recover the feature contexts $\hat{\bs{\mathcal{F}}}_{t}^i$ and pose sequence $\hat{\bs{\mathcal{H}}}_{t}^i$ for masked locations/frames, respectively.

Finally, we fuse multi-level motion features $\boldsymbol{\mathcal{R}}_{t}^{i,P}$ and $\boldsymbol{\mathcal{R}}_{t}^{i,F}$ via an element-wise addition to obtain the final semantic motion representation $\boldsymbol{\mathcal{M}}_{t}^{i}$ that is more robust to pixel degradations:
\begin{equation}
  	\begin{aligned}  
  	\boldsymbol{\mathcal{M}}_{t}^{i} =\boldsymbol{\mathcal{R}}_{t}^{i,P} + \boldsymbol{\mathcal{R}}_{t}^{i,F}.
  	\end{aligned}
  \end{equation}

\subsection{Spatial-Motion Mutual Learning}
\label{sec:SMML}
Fundamentally, the spatial features $\boldsymbol{\mathcal{R}}_t^i$ and the motion features $\boldsymbol{\mathcal{M}}_{t}^{i}$ are complementary and both profitable to the task of VHPE~\cite{cai2020learning, tian2019densely}. Therefore, it would be fruitful to explore how to effectively aggregate them to estimate human poses from videos more accurately. Naively, the motion and spatial features can be aggregated into one feature through convolutions or addition, as done in previous works~\cite{feng2023mutual, pfister2015flowing}. However, such simple aggregation solutions cannot fully exploit both complementary information, leading to suboptimal performance (see Table~\ref{abl-comp}). To address this issue, we propose the Spatial-Motion Mutual Learning (SMML) module that can sufficiently and adaptively fuse spatial and motion cues. The SMML includes three parts: Self-Feature Refinement, Cross-Feature Propagation, and Adaptive Feature Fusion. 

 Given feature sequences $\boldsymbol{\mathcal{R}}_t^i$ and $\boldsymbol{\mathcal{M}}_{t}^{i}$, we aggregate the information of each frame to facilitate subsequent processing. Specifically, we first reshape each frame features within them to 2D feature maps. Then, the features of each frame are concatenated in the channel dimension, and fed into convolutions followed by a flatten operation to obtain aggregated spatial and motion representations $\boldsymbol{\bar{{R}}}_t^i \in \mathbb{R}^{C\times HW}$ and $\bar{\boldsymbol{{M}}}_{t}^{i}\in \mathbb{R}^{C\times HW}$, respectively. The superscript $H$ and $W$ denote the height and width of feature maps.
 
 \begin{figure}[t]
\begin{center}
\includegraphics[width=1.\linewidth]{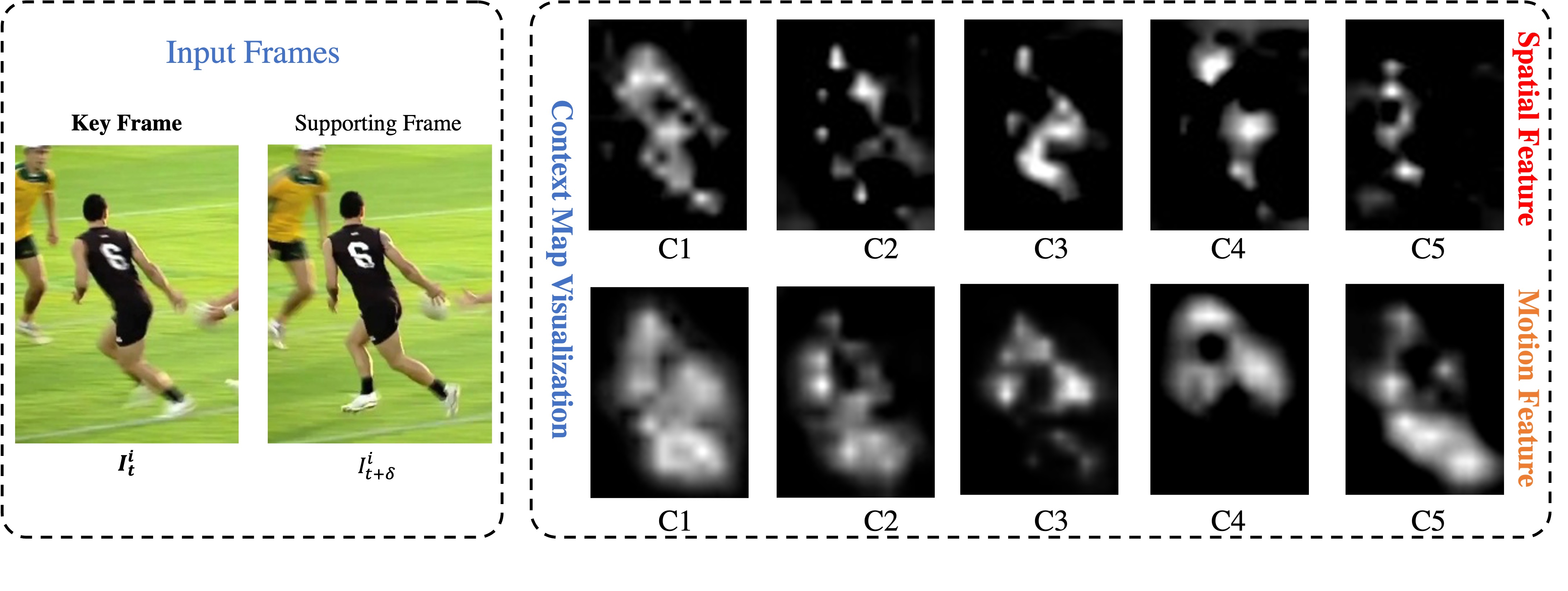}
\end{center}
\caption{\red{\textbf{Visualization of context maps.} The top and bottom rows capture the contexts of spatial ($O_S$) and motion information ($O_M$), respectively. Note that we randomly select five channels $C1-C5$ for visualization.}}
\label{fig:context}
\end{figure}

\noindent
\textbf{Self-Feature Refinement.}\quad Motivated by OCR~\cite{yuan2020object}, we first enhance the feature representations of each modality using the separate context information.  
 \red{Specifically, considering that diverse channels usually contain different semantic contexts, we apply a softmax operation along the channel dimension over  $\boldsymbol{\bar{{R}}}_t^i$ and $\bar{\boldsymbol{{M}}}_{t}^{i}$ to obtain the soft object regions (\emph{i.e.}, context maps) $O \in \mathbb{R}^{C \times HW}$:}
\begin{equation}
  	\begin{aligned}  
  	O_S &= \mathbf{Softmax}\left(\boldsymbol{\bar{{R}}}_t^i\right), \\
  	O_M &= \mathbf{Softmax}\left( \bar{\boldsymbol{{M}}}_{t}^{i} \right).
  	\end{aligned}
  \end{equation}
 \red{Note that we provide several visual samples of context maps $O_S$ and $O_M$ in Fig.~\ref{fig:context}. It is observed that different channels can encode context information of different human body regions.
Then, we compute the context features $OC \in \mathbb{R}^{C \times C}$ as:}
 \begin{equation}
  	\begin{aligned}  
  	OC_S &= \boldsymbol{\bar{{R}}}_t^i \otimes O_S^\mathsf{T},\\
  	OC_M &= \bar{\boldsymbol{{M}}}_{t}^{i} \otimes O_M^\mathsf{T},
  	\end{aligned}
  \end{equation}
where $\otimes$ denotes the matrix multiplication operation. Finally, we calculate the relations between pixels and context features, and employ them to enhance the pixel representations $\boldsymbol{\bar{{R}}}_t^i$ and $\bar{\boldsymbol{{M}}}_{t}^{i}$, obtaining refined counterparts $\boldsymbol{\bar{\bar{{R}}}}_t^i$ and $\bar{\bar{\boldsymbol{{M}}}}_{t}^{i}$, respectively:
\begin{equation}
  	\begin{aligned}  
  	\boldsymbol{\bar{\bar{{R}}}}_t^i &= \mathbf{Conv}\left(\mathbf{Softmax}\left({\boldsymbol{\bar{{R}}}_t^i}^\mathsf{T} \otimes OC_S\right) \otimes OC_S^\mathsf{T}\right),\\
  	\bar{\bar{\boldsymbol{{M}}}}_{t}^{i} &=  \mathbf{Conv}\left(\mathbf{Softmax}\left({\boldsymbol{\bar{{M}}}_t^i}^\mathsf{T} \otimes OC_M\right) \otimes OC_M^\mathsf{T}\right).
  	\end{aligned}
  \end{equation}

\noindent
\textbf{Cross-Feature Propagation.}\quad \red{To fully exploit the complementarity of spatial and motion features, we propose a Spatial-Motion Cross-Attention (SMCA) module which densely transfers the  contexts of each modality to each other. As illustrated in Fig.~\ref{fig:atten} (c), the proposed SMCA involves two inputs, namely a source feature $S$ and a guidance feature $G$.} Specifically, SMCA first applies different convolutions to generate query features $Q$ according to $S$, and key and value features $K$ and $V$ based on $G$. Then, a cross-attention is utilized to sufficiently capture the correlations between source and guidance to propagate the complementary information of guidance features into the source features:
\begin{equation}
  	\begin{aligned}  
  	Atten&(Q,K,V) = \mathbf{Softmax}(QK^\mathsf{T}/\sqrt{d})V,  	\\
    &\tilde{S} = \phi\left(S + Atten(Q,K,V) \right),
  	\end{aligned}
  \end{equation}
where $\phi(\cdot)$ is a convolutional transformation function ($Conv \rightarrow BN \rightarrow ReLU$), and $d$ is a hyperparameter. 

As illustrated in Fig.~\ref{fig:pipeline}, we enforce the spatial feature $\boldsymbol{\bar{\bar{{R}}}}_t^i$ and the motion feature $\bar{\bar{\boldsymbol{{M}}}}_{t}^{i}$ to serve as source and guidance for each other in SMCA, to mutually update themselves:
\begin{equation}
  	\begin{aligned}  
  	\tilde{\boldsymbol{{R}}}_t^i &= \mathbf{SMCA}\left(\boldsymbol{\bar{\bar{{R}}}}_t^i, \bar{\bar{\boldsymbol{{M}}}}_{t}^{i} \right),\\
  	\tilde{\boldsymbol{{M}}}_{t}^{i} &=  \mathbf{SMCA}\left(\bar{\bar{\boldsymbol{{M}}}}_{t}^{i}, \boldsymbol{\bar{\bar{{R}}}}_t^i\right).
  	\end{aligned}
  \end{equation}
By densely propagating the context information between spatial and motion features, both resulted tensors $\tilde{\boldsymbol{{R}}}_t^i$ and $\tilde{\boldsymbol{{M}}}_{t}^{i}$ combine complementary spatial and motion cues.

\noindent
\textbf{Adaptive Feature Fusion.}\quad With preliminarily fused features $\tilde{\boldsymbol{{R}}}_t^i$ and $\tilde{\boldsymbol{{M}}}_{t}^{i}$, we further predict pixel-wise attention weights to adaptively aggregate them together. In particular, we first perform channel concatenation over these two features, and employ a convolutional transformation ($Conv \rightarrow BN \rightarrow ReLU$) to aggregate them, obtaining the tensor $A$. 
\begin{equation}
  	\begin{aligned}
  	A = \mathbf{Conv}\left(\mathbf{Concat}\left(\boldsymbol{\tilde{{R}}}_t^i, \tilde{\boldsymbol{{M}}}_{t}^{i} \right)\right).
   	\end{aligned}
  \end{equation}
Then, $A$ is fed into two separate fully connected (FC) layers, followed by a sigmoid function to predict the attention matrices $A_S$ and $A_M$ for  $\boldsymbol{\tilde{{R}}}_t^i$ and $\tilde{\boldsymbol{{M}}}_{t}^{i}$, respectively. 
\begin{equation}
  	\begin{aligned}
  	A_S &= \mathbf{Sigmoid}\left(\mathbf{FC_S}\left(A\right)\right), \\
  	  	A_M &= \mathbf{Sigmoid}\left(\mathbf{FC_M}\left(A\right)\right).
   	\end{aligned}
  \end{equation}
Finally, we reweight the spatial and motion features to yield the final aggregated representations $F_t^i$:
\begin{equation}
  	\begin{aligned}
  	F_t^i = A_S * \mathbf{Conv}\left(\boldsymbol{\tilde{{R}}}_t^i\right) + A_M * \mathbf{Conv}\left(\tilde{\boldsymbol{{M}}}_{t}^{i}\right).
   	\end{aligned}
  \end{equation}

\noindent
\textbf{Heatmap estimation.}\quad The aggregated feature $F_t^i$ is fed into a detection head ($3 \times 3$ convolution) to obtain the predicted heatmaps $\hat{\mb{H}}_t^{i}$.

\subsection{Training and Inference Algorithms}
 \noindent
\textbf{Training objectives.}\quad
Our training objectives consist of two parts: \textbf{(1)}  We employ the standard pose estimation loss (mean square error) $\mathcal{L}_{H}$ to constrain the training of the final pose estimation:
 \begin{equation}
 	\begin{aligned}
 		\mathcal{L}_{H} =  \left\|\hat{\mb{H}}_t^{i}-\mb{H}_t^i  \right\|_2^2,
 	\end{aligned} 
 \end{equation}
where $\hat{\mb{H}}_t^{i}$ and $\mb{H}_t^i$ symbolize the predicted and ground truth heatmaps, respectively. $\mb{H}_t^i$ is generated via a 2D Gaussian centered at the annotated keypoint locations. \textbf{(2)} A reconstruction loss $\mathcal{L}_{Rec}$ (\emph{i.e.} context and pose reconstruction) is further utilized as an intermediate supervision to facilitate the motion feature learning in MLSME during the training phase:
 \begin{equation}\label{eq:rec}
 	\begin{aligned}
 		\mathcal{L}_{Rec} = \lambda \left\|\hat{\bs{\mathcal{F}}}_{t}^i\hat{M}_c\hat{M}_f - \boldsymbol{\mathcal{R}}_{t}^i\hat{M}_c\hat{M}_f  \right\|_1 + \left\|\hat{\bs{\mathcal{H}}}_{t}^i - \boldsymbol{\mathcal{H}}_{t}^i  \right\|_2^2,
 	\end{aligned} 
 \end{equation}
where $\boldsymbol{\mathcal{H}}_{t}^i$ denotes the ground truth pose heatmaps. Note that we employ the features extracted from the backbone network $\boldsymbol{\mathcal{R}}_{t}^i$ as the context reconstruction target, and only compute the loss for masked locations/frames. The symbol $\lambda$ is a hyperparameter to balance the ratio of different terms. 

Overall, the total loss $\mathcal{L}_{total}$ can be described as:

 \begin{equation}
 	\begin{aligned}
 		\mathcal{L}_{total} = \mathcal{L}_{H} + \mathcal{L}_{Rec}.
 	\end{aligned} 
 \end{equation}

\noindent
\textbf{Inference algorithms.}\quad
During inference, we do not perform any feature masking and employ the MLSME to directly extract multi-level semantic motion features $\boldsymbol{\mathcal{M}}_{t}^{i}$. Then, we aggregate the motion features $\boldsymbol{\mathcal{M}}_{t}^{i}$ and spatial features $\boldsymbol{\mathcal{R}}_t^i$ via SMML to obtain the final pose estimation $\hat{\mb{H}}_t^{i}$.

\renewcommand\arraystretch{1.1}
\begin{table*}[t]
\centering
  \resizebox{1.\textwidth}{!}{
  \begin{tabular}{l|ccccccc|c}
    \hline
      Method &Head   &Shoulder &Elbow       &Wrist   &Hip    &Knee   &Ankle   &{\bf Mean}\cr
      \hline
      PoseTracker~\cite{girdhar2018detect}  &$67.5$ &$70.2$   &$62.0$      &$51.7$  &$60.7$ &$58.7$ &$49.8$  &{$60.6$}\cr
     PoseFlow~\cite{xiu2018pose}         &$66.7$ & $73.3$  &$68.3$      &$61.1$  &$67.5$ &$67.0$ &$61.3$  &{$ 66.5$}\cr
   FastPose~\cite{zhang2019fastpose}  	&$80.0$ &$80.3$   &$69.5$      &$59.1$  &$71.4$ &$67.5$ &$59.4$  &{$ 70.3$}\cr
Simple (R-50)~\cite{xiao2018simple}    &$79.1$ &$80.5$   &$75.5$      &$66.0$  &$70.8$ &$70.0$ &$61.7$  &{$72.4$}\cr
Simple (R-152)~\cite{xiao2018simple}    &$81.7$ &$83.4$   &$80.0$      &$72.4$  &$75.3$ &$74.8$ &$67.1$  &{$ 76.7$}\cr
  STEmbedding~\cite{jin2019multi}        &$83.8$ &$81.6$   &$77.1$      &$70.0$  &$77.4$ &$74.5$ &$70.8$  &{$ 77.0$}\cr
        HRNet~\cite{wang2020deep}        &$82.1$ &$83.6$   &$80.4$      &$73.3$  &$75.5$ &$75.3$ &$68.5$  &{$ 77.3$}\cr
         MDPN~\cite{guo2018multi}       &$85.2$ &$88.5$   &$83.9$      &$77.5$  & $79.0$&$77.0$ &$71.4$  &{$ 80.7$}\cr
   CorrTrack~\cite{rafi2020self}   	  	&$86.1$ &$87.0$   &$83.4$      &$76.4$  & $77.3$&$79.2$ &$73.3$  &{$ 80.8$}\cr 
   Dynamic-GNN~\cite{yang2021learning} 	 &$88.4$ &$88.4$   &$82.0$      &$ 74.5$ &$79.1$ &$78.3$ &$73.1$  &{$81.1$}\cr
   PoseWarper~\cite{bertasius2019learning}  &$81.4$ &$88.3$   &$83.9$      &$ 78.0$ &$82.4$ &$80.5$ &$73.6$  &$81.2$\cr
   DCPose~\cite{liu2021deep} &$ 88.0$  &$ 88.7$     &$ 84.1$   &$78.4$&$ 83.0$        &$ 81.4$&$ 74.2$ &$ 82.8$\cr
   DetTrack~\cite{wang2020combining}  &$89.4$       &$89.7$     &$85.5$ &$79.5$ &$82.4$      &$80.8$       &$76.4$   &$83.8$\cr
    SLT-Pose~\cite{gai2023spatiotemporal} &$88.9$  &$89.7$ &$85.6$ &$79.5$ &$84.2$ &$83.1$ &$75.8$ &$84.2$\cr

    FAMI-Pose~\cite{liu2022temporal} 	&$ 89.6$  &$ 90.1$ &$ 86.3$ &$80.0$ &$ 84.6$ &$83.4$ &$ 77.0$ &$ 84.8$\cr
    TDMI~\cite{feng2023mutual} 	&$ 90.0$  &$ 91.1$ &$ 87.1$ &$ 81.4$ &$ 85.2$ &$  84.5$ &$ 78.5$ &$ 85.7$\cr
	   DSTA~\cite{he2024video}  &$ 89.3$  &$ 90.6$ &$ 87.3$ &$ 82.6$ &$ 84.5$ &$  85.1$ &$ 77.8$ &$ 85.6$\cr   
\hline
      \bf SDTC (Ours) 	&$\bf 90.1$  &$\bf 92.1$ &$ \bf 89.1$ &$\bf 85.1$ &$\bf 86.3$ &$\bf  87.7$ &$\bf 81.9$ &$\bf 87.5$\cr 
    \hline
    \end{tabular}}
      \caption{\textbf{Quantitative comparisons} on the PoseTrack2017 validation set.}  \label{17val}
    \end{table*}

 \renewcommand\arraystretch{1.1}
\begin{table*}[t]
\centering
   \resizebox{1.\textwidth}{!}{
   \begin{tabular}{l|ccccccc|c}
     \hline
      Method                         &Head &Shoulder &Elbow  &Wrist &Hip &Knee &Ankle &{\bf Mean}\cr
     \hline
 AlphaPose~\cite{fang2017rmpe}        &$63.9$  &$78.7$&$77.4$ &$71.0$ &$73.7$ &$73.0$    &$69.7$     &{$71.9$}\cr
 MDPN~\cite{guo2018multi}               &$75.4$ &$81.2$ &$79.0$ &$74.1$ &$72.4$ &$73.0$  &$69.9$   &{$75.0$}\cr
 PGPT~\cite{bao2020pose}    	 		&-       &-     &-      &$72.3$ &-      &-       &$72.2$   &{$76.8$}\cr
 Dynamic-GNN~\cite{yang2021learning}  	&$80.6$ &$84.5$   &$80.6$  &$ 74.4$ &$75.0$ &$76.7$ &$71.8$  &$77.9$\cr
 PoseWarper~\cite{bertasius2019learning}  &$79.9$&$86.3$&$82.4$&$77.5$&$79.8$&$78.8$&$73.2$  &$79.7$\cr
 PT-CPN++~\cite{yu2018multi}	 &$82.4$ &$88.8$ &$86.2$ &$79.4$ &$72.0$ &$80.6$ &$76.2$  &$80.9$\cr
 DCPose~\cite{liu2021deep} 		 &$ 84.0$ &$ 86.6$&$ 82.7$&$ 78.0$&$ 80.4$&$ 79.3$&$ 73.8$&$ 80.9$\cr 
 DetTrack~\cite{wang2020combining} &$84.9$ &$87.4$ &$84.8$ &$79.2$ &$77.6$      &$79.7$       &$75.3$   &$81.5$ \cr
  SLT-Pose \cite{gai2023spatiotemporal}  &$ 84.3$&$ 87.5$&$83.5$&$ 78.5$&$ 80.9$&$ 80.2$&$ 74.4$&$ 81.5$\cr
 FAMI-Pose~\cite{liu2022temporal} &$ 85.5$&$ 87.7$&$ 84.2$&$ 79.2$&$ 81.4$&$81.1$&$ 74.9$&$ 82.2$\cr
 TDMI~\cite{feng2023mutual}	 &$\bf 86.2$&$ 88.7$&$ 85.4$&$ 80.6$&$\bf 82.4$&$ 82.1$&$ 77.5$&$ 83.5$\cr
 DSTA~\cite{he2024video}  &$ 85.9$  &$ \bf88.8$ &$ 85.0$ &$ 81.1$ &$ 81.5$ &$  83.0$ &$ 77.4$ &$ 83.4$\cr 
\hline
    \bf SDTC (Ours) 	  &$ 84.9$  &$ 88.6$ &$\bf 86.1$ &$\bf 83.1$ &$ 82.3$ &$\bf 85.2$ &$\bf 80.7$ &$\bf 84.3$\cr
        \hline
     \end{tabular}}
     \caption{\textbf{Quantitative results} on the PoseTrack2018 validation set.}  \label{18val}
   \end{table*}

\renewcommand\arraystretch{1.1}
\begin{table*}[t]
\centering
   \resizebox{1.\textwidth}{!}{
   \begin{tabular}{l|ccccccc|c}
     \hline
     Method &Head &Shoulder &Elbow  &Wrist &Hip &Knee &Ankle &{\bf Mean}\cr
     \hline
SimpleBaseline~\cite{xiao2018simple} 	  &$80.5$ &$81.2$ &$73.2$ &$64.8$ &$73.9$ &$72.7$  &$67.7$   &$73.9$\cr
HRNet~\cite{wang2020deep}           	    &$81.5$ &$83.2$ &$81.1$ &$75.4$ &$79.2$ &$77.8$  &$71.9$   &$78.8$\cr
 PoseWarper~\cite{bertasius2019learning}    &$82.3$ &$84.0$ &$82.2$ &$75.5$ &$80.7$ &$78.7$  &$71.6$   &$79.5$\cr
DCPose~\cite{liu2021deep}				  &$ 83.2$&$ 84.7$&$ 82.3$&$ 78.1$&$ 80.3$&$ 79.2$&$ 73.5$&$ 80.5$\cr 
 FAMI-Pose~\cite{liu2022temporal} 		    &$ 83.3$&$ 85.4$&$ 82.9$&$ 78.6$&$ 81.3$&$80.5$&$ 75.3$&$ 81.2$\cr
 SLT-Pose \cite{gai2023spatiotemporal}   &$ 83.3$&$ 85.1$&$82.7$&$ 78.5$&$ 81.3$&$ 80.8$&$ 75.6$&$ 81.3$\cr
 TDMI~\cite{feng2023mutual} 			 	  &$ 85.8$&$\bf 87.5$&$ 85.1$&$ 81.2$&$ 83.5$&$82.4$&$ 77.9$&$ 83.5$\cr
 DSTA~\cite{he2024video} 					  &$\bf 87.5$&$ 87.0$ &$ 84.2$ &$ 81.4$ &$ 82.3$ &$  82.5$ &$ 77.7$ &$ 83.5$\cr 
\hline
      \bf SDTC (Ours)	&$ 86.0$  &$ 87.3$ &$\bf 86.2$ &$\bf 84.0$ &$\bf 83.7$ &$\bf  85.1$ &$\bf 81.1$ &$\bf 84.9$\cr
      \hline
     \end{tabular}}
     \caption{\textbf{Quantitative results} on the PoseTrack21 dataset. } \label{21val}
   \end{table*}  

 \begin{figure*}[t]
\begin{center}
\includegraphics[width=1.\linewidth]{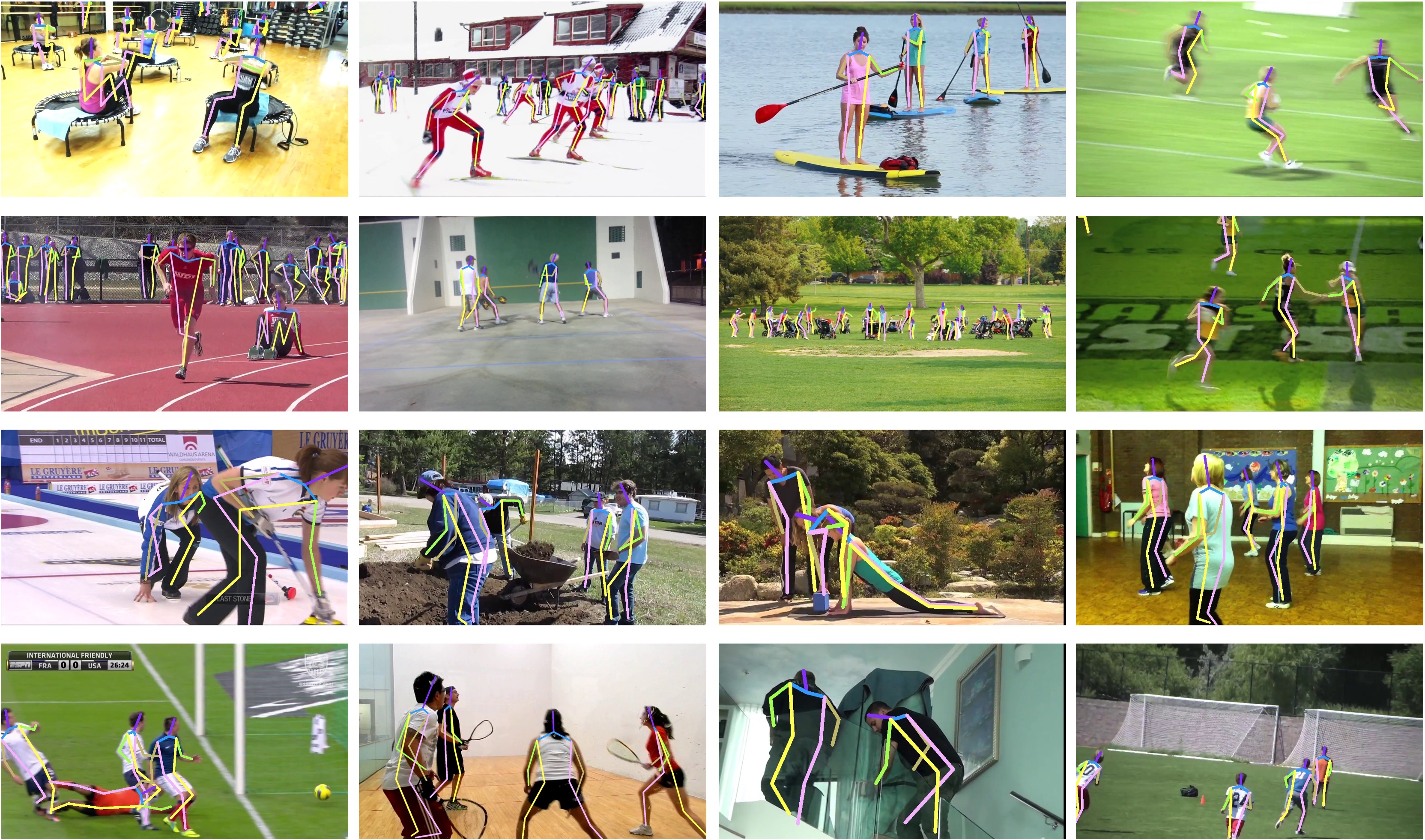}
\end{center}
\caption{\textbf{Qualitative examples} of SDTC on benchmark datasets. Challenging scenes \emph{e.g.} occlusion and blur are involved.}
\label{fig:results}
\end{figure*}

\section{Experiments}
\label{sec:experiments}
\subsection{Experimental Settings}
 
  \noindent
\textbf{Datasets.}\quad
\red{We evaluate the proposed SDTC on PoseTrack~\cite{Iqbal_2017_CVPR, Andriluka_2018_CVPR, doering2022posetrack21}, a series of large-scale benchmark datasets for video-based human pose estimation that contain challenging sequences of highly cluttered people performing various rapid movements.} Specifically, \textbf{PoseTrack2017}~\cite{Iqbal_2017_CVPR} includes $250$ videos for training and $50$ videos for validation, with a total of $80,144$ pose annotations. \textbf{PoseTrack2018}~\cite{Andriluka_2018_CVPR} increases the number of videos to $593$ for training and $170$ for validation, and provides $153,615$ human pose labels. Both datasets annotate $15$ anatomical keypoints and contain an extra flag for visibility. \textbf{PoseTrack21}~\cite{doering2022posetrack21} further augments the PoseTrack2018 dataset, especially for pose annotations of particular small persons and persons in crowds, including $177,164$ pose labels. The flag of the joint visibility is re-defined to indicate the occlusion cases.
 
  \noindent
 \textbf{Evaluation metric.}\quad 
\red{Following previous works~\cite{wang2020deep, xiao2018simple}, we employ the metric of average precision (\textbf{AP}) to evaluate our model. We first compute the AP for each joint and then obtain the final performance (\textbf{mAP}) by averaging over all joints.}
  
 \noindent
 \textbf{Implementation details.}\quad 
Our framework is implemented with PyTorch~\cite{paszke2019pytorch}. \red{During training, we incorporate data augmentation strategies including random scaling $[0.65, 1.35]$,  random rotation $[-45^\circ, 45^\circ]$, truncation, and flipping.} The input image size is set to $256 \times 192$. \red{The temporal span $\delta$ is set to $2$.} To weight different loss terms in Eq.~\ref{eq:rec}, we empirically set $\lambda=0.01$. We employ the AdamW optimizer with a base learning rate of $5e-4$ (decays to $5e-5$ and $5e-6$ at the $20$-th and $40$-th epochs, respectively). All training processes are performed on a TITAN RTX GPU and terminated within $50$ epochs.

 \subsection{Comparison with State-of-the-art Approaches}

 \noindent
 \textbf{Results on the PoseTrack2017 dataset.}\quad
 We first benchmark the proposed SDTC on the PoseTrack2017 validation set. A total of $18$ models are compared, and the experimental results are tabulated in Table \ref{17val}. From this table, we can observe that our proposed SDTC outperforms previous state-of-the-art methods over all joints, achieving the final performance of $87.5$ mAP. Compared to the well-established approaches TDMI~\cite{feng2023mutual} and DSTA~\cite{he2024video}, SDTC delivers a remarkable performance gain of $1.8$ mAP and $1.9$ mAP, respectively. The performance improvement for challenging joints is also encouraging: we attain an mAP of $85.1$ ($\uparrow 2.5$) for wrists and $81.9$ ($\uparrow 4.1$) for ankles. 
 \red{Such remarkable and consistent performance boost demonstrates the importance of explicitly embracing semantical motion information and fully aggregating motion and spatial features.}  Moreover, we display example visualizations for challenging scenes including mutual occlusion and fast motion in Fig.~\ref{fig:results}, which attest to the robustness of our method.

 \noindent
 \textbf{Results on the PoseTrack2018 dataset.}\quad
 Table \ref{18val} reports the results of our method as well as existing state-of-the-art approaches on the PoseTrack2018 validation set. The proposed SDTC delivers an mAP of $84.3$, which once again surpasses other approaches. SDTC reaches the final accuracy of $86.1$ mAP, $83.1$ mAP, $85.2$ mAP, and $80.7$ mAP for elbow, wrist, knee, and ankle joints, respectively.

 \noindent
 \textbf{Results on the PoseTrack21 dataset.}\quad
 Furthermore, we evaluate our proposed method on the PoseTrack21 dataset. Detailed comparisons are provided in Table \ref{21val}. We observe that the previous method~\cite{he2024video} has already yielded an impressive performance. In contrast, our approach can obtain $84.9$ ($\uparrow 1.4$) mAP. On the other hand, compared to the pose estimation results on PoseTrack2018, SDTC achieves a better performance in more challenging PoseTrack21 ($84.3$ mAP v.s. $84.9$ mAP). This might be evidence to show the merit of our approach especially for challenging cases.

 \renewcommand\arraystretch{1.1}
\begin{table*}[t]
\centering
   \resizebox{0.9\textwidth}{!}{
   \begin{tabular}{l|cc|ccc|c}
    \hline
    Method&\bf MLSME &Optical Flow   &\bf SMML &Add &Conv.  &Mean\cr
    \hline
    (a) Optical-Add. & &\checkmark  &  &\checkmark & & $84.0$\cr
    (b) Optical-Conv. & &\checkmark  &  & &\checkmark & $83.9$\cr
     \hline
    (c) MLSME-Add. &\checkmark &  &  &\checkmark & & $86.2$\cr
    (d) MLSME-Conv. &\checkmark &  &  & &\checkmark & $86.0$\cr
     \hline
    \bf STDC &\checkmark &  & \checkmark & &  & $\bf 87.5$\cr
    \hline
    \end{tabular}}
   \caption{{Ablation of different components (\textbf{MLSME} and \textbf{SMML})}.} 
   \label{abl-comp} 
   \end{table*}     
 
  \renewcommand\arraystretch{1.1}
\begin{table*}[!t]
\centering
   \resizebox{0.9\textwidth}{!}{
   \begin{tabular}{l|cc|c|c}
    \hline
   Method & Patch-level motion. &Frame-level motion. &Masking ratio $r$ &Mean\cr
    \hline
    (a) & & &-  & $85.6$\cr
    \hline
    (b) &\checkmark & &$50\%$  & $86.7$\cr
    \bf STDC &\checkmark &\checkmark &$50\%$  & $\bf 87.5$\cr
    \hline
    (c) &\checkmark &\checkmark &$25\%$  & $86.9$\cr
    (d) &\checkmark &\checkmark &$75\%$  & $87.4$\cr
    \hline
    \end{tabular}}
   \caption{Ablation on \textbf{Multi-Level Semantic Motion Encoder (MLSME)}.} 
   \label{abl-mlsme} 
   \end{table*}

\begin{figure}[t]
\begin{center}
\includegraphics[width=0.7\linewidth]{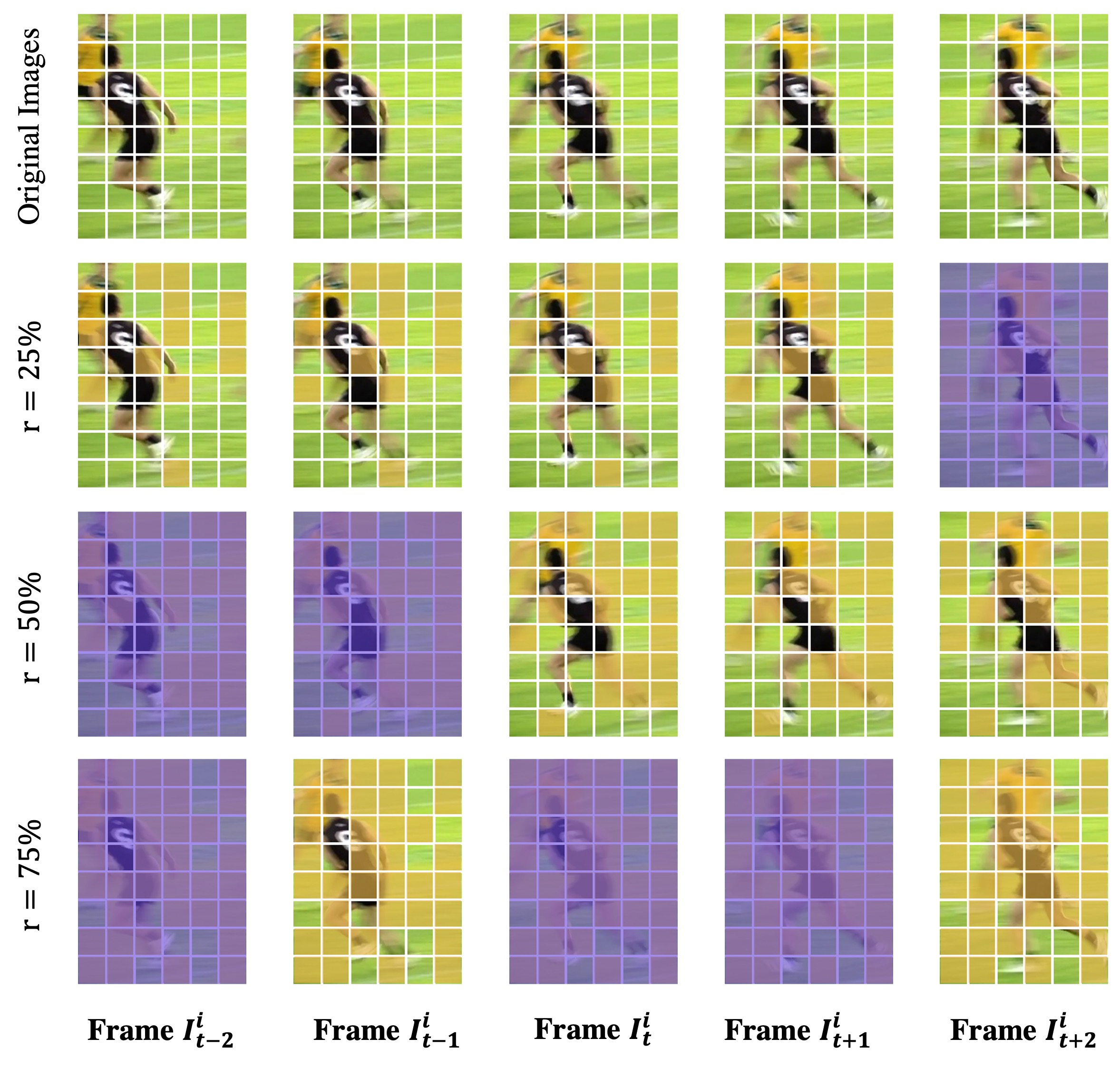}
\end{center}
\caption{\red{\textbf{Visualization of masking ratio $r$.} The patches/frames in yellow and blue indicate the random tube masking and frame masking, respectively.}}
\label{fig:mask}
\end{figure}

   \renewcommand\arraystretch{1.1}
\begin{table*}[t]
\centering
   \resizebox{0.9\textwidth}{!}{
   \begin{tabular}{l|ccc|c}
    \hline
   Method & Self refinement &Cross propagation &Adaptive fusion &Mean\cr
    \hline
    (a) &\checkmark & &  & $86.6$\cr
    (b) &\checkmark &\checkmark &  & $87.2$\cr
    \bf STDC &\checkmark &\checkmark &\checkmark  & $\bf 87.5$\cr
    \hline
    \end{tabular}}
   \caption{Ablation on \textbf{Spatial-Motion Mutual Learning (SMML)}.} 
   \label{abl-smml} 
   \end{table*}

   \renewcommand\arraystretch{1.}
\begin{table*}[!h]
\centering
   \resizebox{0.8\textwidth}{!}{
   \begin{tabular}{l|cc|c}
    \hline
   Method & Spatial encodings &Temporal  encodings &Mean\cr
    \hline
    \red{(a)} & &  & \red{$86.5$}\cr
    (b) &\checkmark & &  $87.1$\cr
    \bf STDC &\checkmark &\checkmark  & $\bf 87.5$\cr
    \hline
    \end{tabular}}
   \caption{\red{Ablation on \textbf{positional encodings}.}} 
   \label{abl-pos} 
   \end{table*}  
   
\renewcommand\arraystretch{1.1}
\begin{table*}[!h]
\centering
   \resizebox{0.8\textwidth}{!}{
   \begin{tabular}{c|c|c|c}
     \hline
        Parameter  &$\lambda=0.01$  &$ {\lambda = 0.1}$ &$\lambda=1$\cr
    \hline
   Mean Accuracy (mAP) &$\bf 87.5$  & $87.3$  & $87.2$\cr
	 \hline
     \end{tabular}}
     \caption{{Ablation of modifying the  \textbf{loss ratio $\lambda$}.}} 
      \label{abl-alpha}
   \end{table*}

   \renewcommand\arraystretch{1.1}
\begin{table*}[t]
\centering
   \resizebox{0.8\textwidth}{!}{
   \begin{tabular}{c|c|c}
     \hline
       Methods   &Temporal span $\delta$ &Mean\cr
    \hline
    $2$ supporting frames, $\{-1,1\}$ &$\delta=1$  & $86.8$\cr
    $4$ supporting frames, $\{-2,-1,1,2\}$ &$\delta=2$  & $\bf 87.5$\cr
    $6$ supporting frames, $\{-3,-2,-1,1,2,3\}$ &$\delta=3$  & $\bf 87.5$\cr
	 \hline
     \end{tabular}}
     \caption{{Ablation of modifying the \textbf{temporal span $\delta$}.}} 
      \label{abl-delta}
   \end{table*}

  \renewcommand\arraystretch{1.}
\begin{table*}[!h]
\centering
   \resizebox{0.7\textwidth}{!}{
   \begin{tabular}{l|cc|c}
     \hline
     Method  &\#Params.  &GFLOPs &Performance\cr
     \hline
 
 PoseWarper~\cite{bertasius2019learning}  &$ 7.5$ M &$210.5$&$ 81.2$\cr
  SLT-Pose \cite{gai2023spatiotemporal}   &$23.1$M &$320.6$ &$84.2$\cr
 TDMI~\cite{feng2023mutual}		  &$\bf 5.3$ M &$198$&$ 85.7$\cr
 
 \hline
 \bf SDTC  &$ 10.4$ M &$\bf 177$ &$\bf 87.5$\cr 
     \hline
     \end{tabular}}
     \caption{\textbf{Computation complexity} of different methods.} \label{run}
   \end{table*}

\subsection{Ablation Study}
We conduct ablation studies to examine the contribution of proposed components and design choices. All experiments are performed on PoseTrack2017. 

\noindent
\textbf{Study on components.}\quad
We evaluate the efficacy of our proposed components, including the Multi-Level Semantic Motion Encoder (MLSME) and the Spatial-Motion Mutual Learning (SMML), and provide the results in Table~\ref{abl-comp}. 
\textbf{(1)} We first construct two baselines, \textbf{(a)} Optical-Add. and \textbf{(b)} Optical-Conv., which employ optical flow as pixel-wise motion features, and fuse spatial and motion features via element-wise addition and convolutions, respectively. These two baselines produce performances of $84.0$ mAP and $83.9$ mAP. \textbf{(2)} Then, we remove optical flow and incorporate the MLSME into baselines \textbf{(a)} and \textbf{(b)} for extracting semantical motions, forming methods \textbf{(c)} MLSME-Add. and \textbf{(d)} MLSME-Conv., respectively. The mAP increases from $84.0$ and $83.9$ to $86.2$ ($\uparrow 2.2$) for \textbf{(c)} and $86.0$ ($\uparrow 2.1$) for \textbf{(d)}. Such significant performance improvements corroborate the effectiveness of our MLSME in introducing semantical motion information to facilitate the task of video-based human pose estimation. These experiments also suggest that our MLSME can derive more robust motion representations compared to the optical flow. \textbf{(3)} Our complete SDTC further introduces SMML to fully aggregate spatial and motion features, achieving the best performance of $87.5$ mAP. Compared to simple feature aggregation schemes such as element-wise addition or concatenation\&convolution, our SMML can improve the mAP by $1.3$ and $1.5$. This highlights the superiority of SMML in taking full advantage of spatial and motion cues.

\noindent
\textbf{Multi-Level Semantic Motion Encoder (MLSME).}\quad
In this ablation setting, we examine the effects of various specific designs in the Multi-Level Semantic Motion Encoder (MLSME). Experimental results are provided in Table~\ref{abl-mlsme}. \textbf{(a)} We first remove the patch- and frame-level motion encoders (\emph{i.e.} masked reconstruction strategy), and employ plain vision transformers to extract motion features. This baseline reduces the performance by $1.9$ mAP. \textbf{(b)} Next, we incorporate the temporal tube masking and the patch-level motion encoder, exploring patch-level spatiotemporal semantical correlations  among frames to extract motion features. This module significantly increases the mAP by $1.1$. By further introducing the random frame masking and the frame-level motion encoder, our SDTC can extract multi-level semantic motion features which obtains the best performance ($\uparrow 0.8$ mAP). These results demonstrate the effectiveness of the proposed multi-masked context and pose reconstruction strategy in deriving more robust motion representations. 

We point out that the proposed patch- and frame-level motion encoders adopt a same masking ratio during training. \red{We further study the effects of the masking ratio $r$ over pose estimation performance. Three experiments are conducted, in which the masking ratio is set to $r = 25\%$, $r = 50\%$, and $r = 75\%$. Example visualizations of masking ratio $r$ in original frames are depicted in Fig.~\ref{fig:mask} for better viewing.} From the results in Table~\ref{abl-mlsme}, we observe that $r=50\%$ is the most effective and we take this as the default setting.

\noindent
\textbf{Spatial-Motion Mutual Learning (SMML).}\quad
Moreover, we study the impact of various micro designs within the Spatial-Motion Mutual Learning (SMML), including the self-feature refinement, cross-feature propagation, and the adaptive feature fusion. \textbf{(a)} To aggregate spatial and motion features, we perform self refinement to enhance them separately as described in Sec.~\ref{sec:SMML}, and then add them to obtain the fused features. From the results in Table~\ref{abl-smml}, we can observe that this baseline yields an mAP of $86.6$. Compared to the simple scheme that directly performing addition over spatial and motion features (Table~\ref{abl-comp} (c)), this method produces a performance boost of $0.4$ mAP. This shows the important role of feature enhancement using corresponding context information.  \textbf{(b)} By incorporating the cross-feature propagation operation into \textbf{(a)} to fully discover complementary information for each other, the performance is significantly improved to $87.2$ mAP ($\uparrow 0.6$). \textbf{(c)} Finally, our complete framework further generates pixel-wise attention weights to adaptively aggregate spatial and motion features, delivering the best performance ($87.5$ mAP).

\noindent
\textbf{Positional encodings.}\quad
\red{In addition, we adopt different types of positional encodings to examine their influence on the final performance, and tabulate the results in Table \ref{abl-pos}. For the first setting \textbf{(a)}, we remove both spatial and temporal encodings, and do not leverage any positional embeddings. This baseline yields an mAP of $86.5$. \textbf{(b)} Next, we introduce spatial encodings to indicate space locations for each token which increases the mAP to $87.1$. Our complete SDTC further incorporates temporal encodings over \textbf{(b)} to indicate time locations, delivering the best performance of $87.5$ mAP ($\uparrow 0.4$). Such experimental results demonstrate the important roles of both spatial and temporal encodings in spatiotemporal modeling.}

   \begin{figure*}[t]
\begin{center}
\includegraphics[width=1.\linewidth]{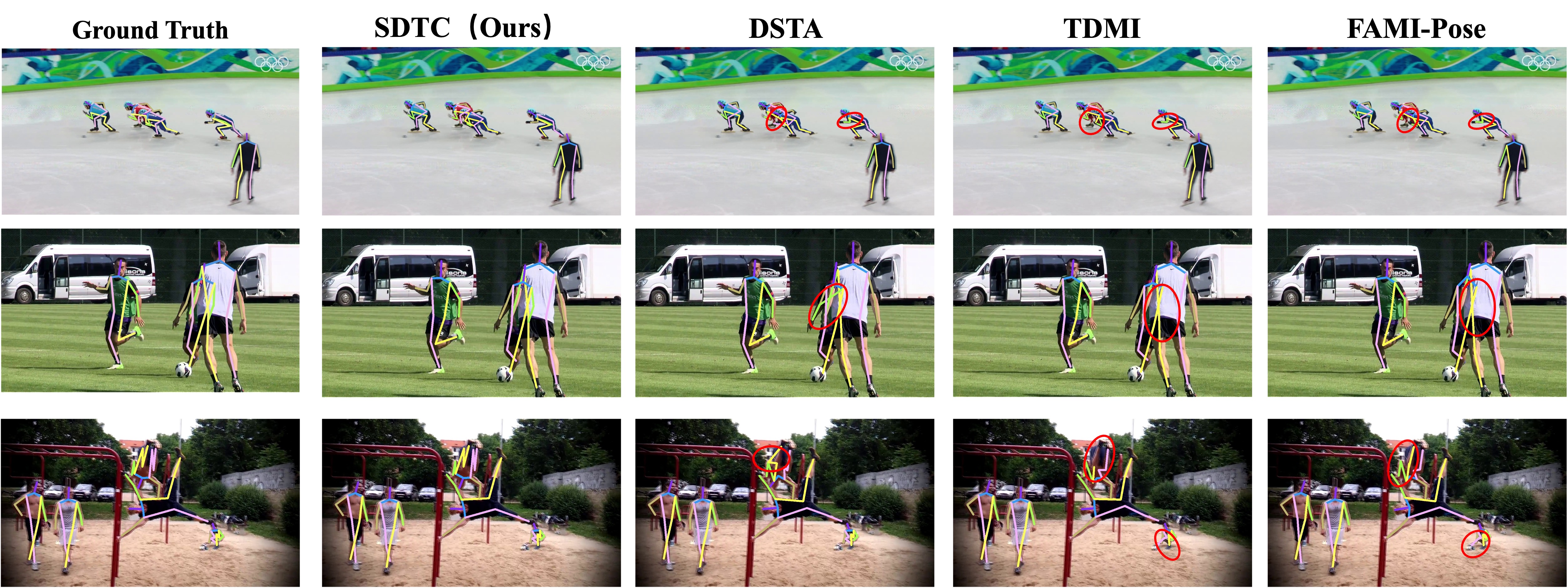}
\end{center}
\caption{\red{\textbf{Visual comparisons} of different approaches on the PoseTrack dataset. Inaccurate keypoint detections are highlighted by red circles. The ground truth human poses are included in the first column.}}
\label{fig:vis}
\end{figure*} 

\noindent
\textbf{Loss ratio $\lambda$.}\quad
Recall that we use $\lambda$ to balance the training of context and pose reconstruction in Eq.~\ref{eq:rec}. We examine the influence of modifying different $\lambda$ and report the results in Table \ref{abl-alpha}. We empirically observe that in the setting of $\lambda=0.01$, the context and pose reconstruction losses are numerically well-balanced which delivers the best performance. When increasing the ratio of context reconstruction objective, the performance slightly decreases by $0.2$ mAP for $\lambda = 0.1$ and $0.3$ mAP for $\lambda = 1$, respectively.

\noindent
\textbf{Temporal span $\delta$.}\quad
Furthermore, we study the effects of adopting different temporal span $\delta$ that controls the number of supporting frames. The results in Table \ref{abl-delta} reflect a gradual performance improvement with  increasing $\delta$, whereby the mAP increases from $86.8$ for $\delta=1$ to $87.5$, $87.5$ at $\delta=2$ and $\delta=3$, respectively. This is in accordance with our expectation, \emph{i.e.}, incorporating more supporting frames enables accessing larger temporal contexts, which facilitates more accurate pose estimation. Another observation is that the pose estimation performance saturates from $\delta=2$. This might be attributed to the fact that the performance boost gained from the temporal information has been gradually saturated.

\noindent
\textbf{Computation complexity.}\quad We perform the computational cost comparisons of our SDTC with existing video-based human pose estimation methods in Table~\ref{run}. It is observed that our SDTC achieves a better tradeoff between computational cost and performance. Compared to PoseWarper~\cite{bertasius2019learning} and TDMI~\cite{feng2023mutual}, SDTC delivers a better performance ($87.5$ mAP) with a similar magnitude of trainable model parameters ($10.4$M) and fewer GFLOPs ($177$).

 \begin{figure*}[!t]
\begin{center}
\includegraphics[width=.8\linewidth]{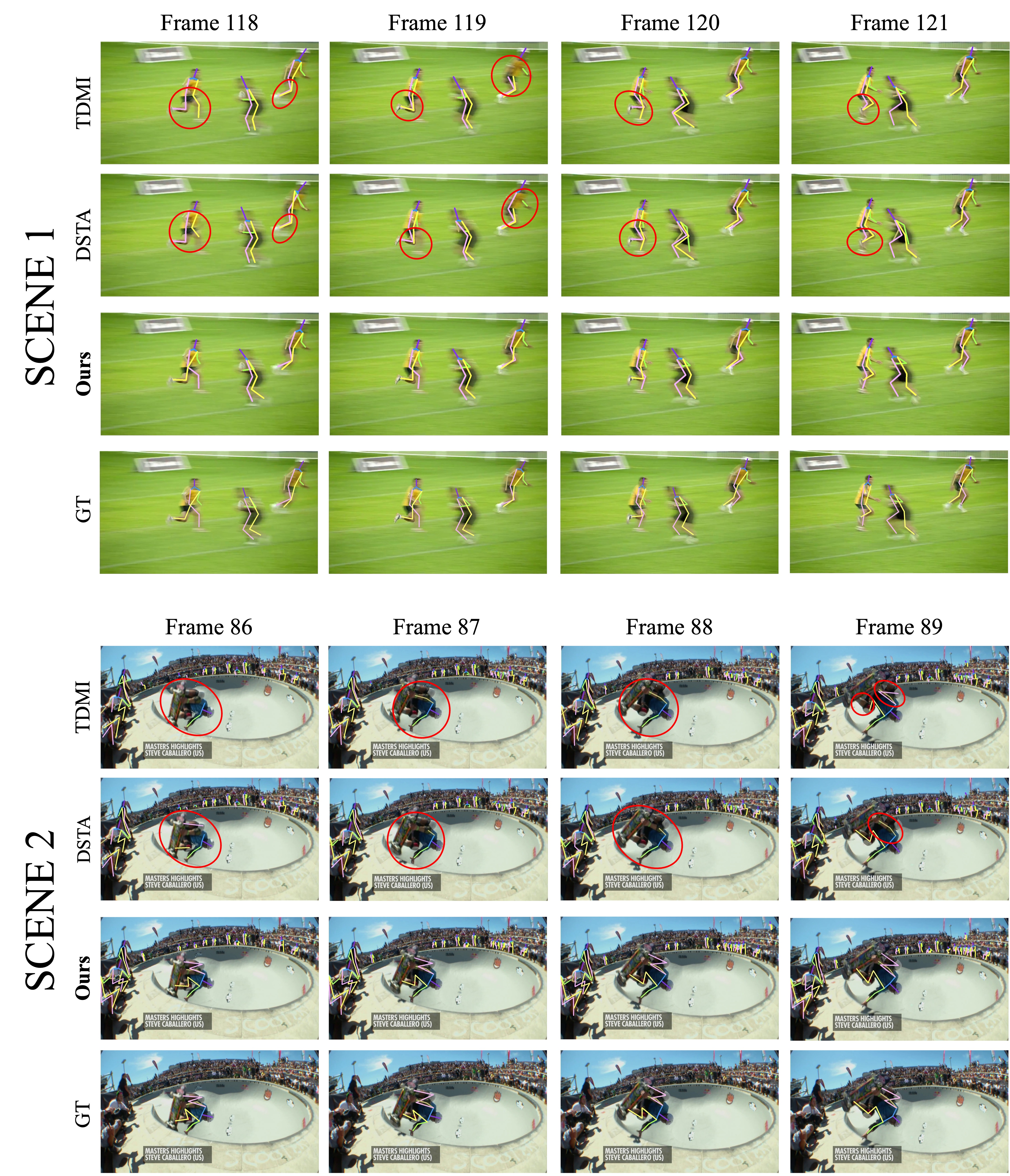}
\end{center}
\caption{\red{\textbf{Visual comparisons} of the pose estimations of SDTC (Ours), DSTA, and TDMI on challenging sequences from the PoseTrack dataset. Inaccurate detections are highlighted by red circles. The ground truth human poses are provided in the last row.}}
\label{fig:com}
\end{figure*}

\begin{figure}[t]
\begin{center}
\includegraphics[width=0.8\linewidth]{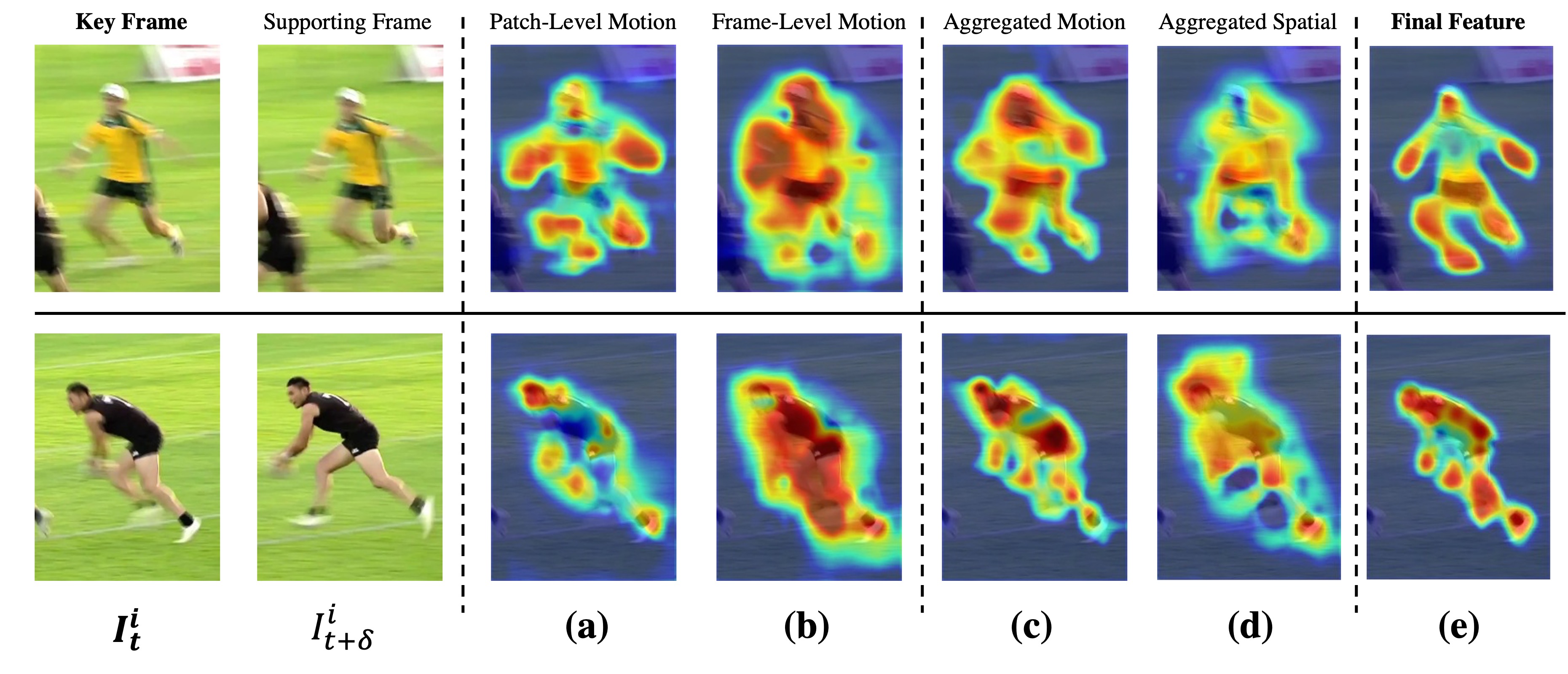}
\end{center}
\vspace{-1em}
\caption{\red{Visualization of various \textbf{intermediate feature representations.}}}
\label{fig:feature}
\end{figure}

\begin{figure}[!h]
\begin{center}
\includegraphics[width=0.8\linewidth]{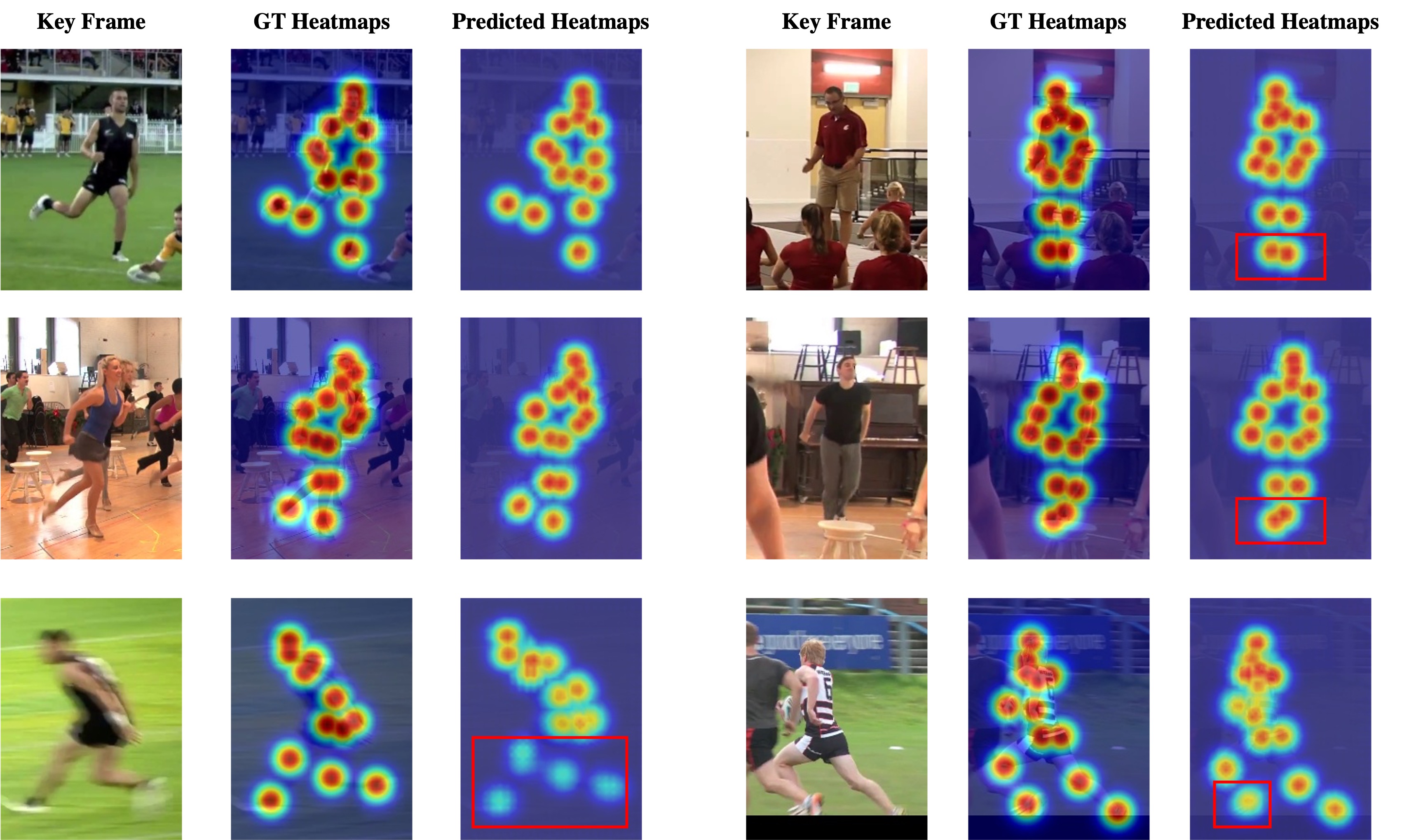}
\end{center}
\vspace{-1em}
\caption{\red{Visualization of our predicted \textbf{pose heatmaps}  and the corresponding ground truth counterparts. Challenging cases such as occlusion or blur are highlighted by red rectangles.}}
\label{fig:heatmap}
\end{figure}

  \begin{figure*}[h]
\begin{center}
\includegraphics[width=.96\linewidth]{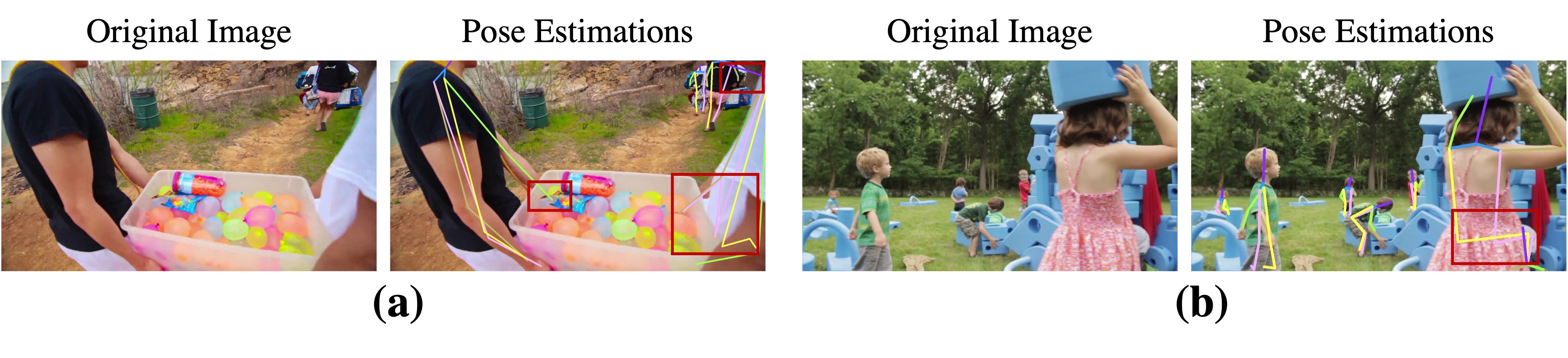}
\end{center}
\vspace{-1em}
\caption{\textbf{Example visualizations of failed pose estimations}. Inaccurate detections are highlighted by rectangles.}
\label{limit}
\end{figure*}

\subsection{Qualitative Analysis}
\red{In addition to the quantitative results, we also conduct extensive qualitative analyses of the proposed method, including comparison of visual results, representation visualization, heatmap visualization, and limitations (failure cases).}

 \noindent
 \textbf{Comparison of visual results.}\quad
\red{We first qualitatively examine the ability of our model to handle challenging cases such as occlusions and blur. We display in Fig.~\ref{fig:vis} the side-by-side comparisons of SDTC against state-of-the-art VHPE methods DSTA~\cite{he2024video}, TDMI~\cite{feng2023mutual} and FAMI-Pose~\cite{liu2022temporal}. Note that we also provide corresponding ground truth human poses in the first column for easier comparisons. Existing methods struggle to explore rich semantic correlations across frames and adequate spatial-motion feature aggregation, resulting in suboptimal performance.} Through the principled design of MLSME and SMML, our approach shows a better ability to deal with complex cases. \red{Moreover, we illustrate sequential comparisons of SDTC against TDMI and DSTA in Fig.~\ref{fig:com}.} This further demonstrates the effectiveness of our method.
 
\noindent
\red{\textbf{Representation visualization.}\quad To better understand the mechanism behind the proposed method, we further provide the visualizations of various intermediate feature representations, including (a) patch-level motion feature $\boldsymbol{\mathcal{R}}_{t}^{i,P}$, (b) frame-level motion feature $\boldsymbol{\mathcal{R}}_{t}^{i,F}$, (c) aggregated motion feature $\bar{\boldsymbol{{M}}}_{t}^{i}$, (d) aggregated spatial feature $\boldsymbol{\bar{{R}}}_t^i$, and (e) the final feature $F_t^i$. All visual samples are depicted in Fig.\ref{fig:feature}. From this figure, we can observe that: \textbf{(1)} The patch-level  and frame-level motion features (\emph{i.e.},  $\boldsymbol{\mathcal{R}}_{t}^{i,P}$ and $\boldsymbol{\mathcal{R}}_{t}^{i,F}$) exhibit distinct characteristics, where the former scatters across significant local human parts while the later attends to global information. This is in line with our intuitions on Patch- and Frame-Level Motion Encoder. \textbf{(2)} The aggregated motion feature $\bar{\boldsymbol{{M}}}_{t}^{i}$ and spatial feature $\boldsymbol{\bar{{R}}}_t^i$ are complementary to each other, and both of them are valuable for pose estimation. 
This corroborates our motivation for designing SMML to take full advantage of spatial and motion representations. \textbf{(3)} The final fused feature $F_t^i$ (derived from SMML) is more compact and delicate, which is beneficial for accurate pose estimation.
}

\noindent
\red{\textbf{Heatmap visualization.}\quad Moreover, we illustrate in Fig.~\ref{fig:heatmap} the predicted pose heatmaps of our method in different scenarios. Note that we provide the ground truth heatmaps for comparison. It is observed that our approach can produce robust heatmap predictions across various cases, including occlusion or motion blur.
 }
 
\noindent
 \textbf{Limitations.}\quad Visualized results show that our approach can achieve robust pose estimations in challenging cases. However, the proposed SDTC still may fail when the human body in the frame is highly incomplete (\emph{i.e.}, containing only a small number of visible joints). As illustrated in Fig.~\ref{limit}, for persons who are close to the camera, the model often fails to fully understand the human body information and produces inaccurate joint detections.
 
 \section{Conclusion and Future Works} 
 In this paper, we propose a novel approach which explores robust multi-level semantical motion modeling and dense spatio-temporal collaboration for video-based human pose estimation. We design a Multi-Level Semantic Motion Encoder to acquire motion dynamics that are insensitive to pixel degradations by fully learning multi-level semantic relationships among frames. We further introduce a Spatial-Motion Mutual Learning module, densely propagating and consolidating complementary contexts to enhance spatial-motion feature aggregation. Extensive experiments show that our approach achieves state-of-the-art performance on three large-scale benchmark datasets, PoseTrack2017, PoseTrack2018, and PoseTrack21. Future works include diverse applications to other vision tasks such as 3D human pose estimation and pose tracking.

\bibliographystyle{elsarticle-num} 
\bibliography{References}

\newpage

\subsection*{  }
\setlength\intextsep{0pt} 
\begin{wrapfigure}{l}{25mm}
    \centering
    \includegraphics[width=1in,height=1.25in,clip,keepaspectratio]{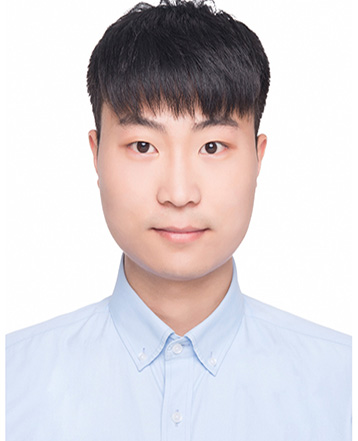}
\end{wrapfigure}
\noindent \textbf{Runyang Feng} is a Ph.D. student in the School of Artificial Intelligence at Jilin University. His current research focuses on computer vision, 2D human pose estimation, and video representation learning.\par

\hspace*{\fill}

\subsection*{  }  
\setlength\intextsep{0pt} 
\begin{wrapfigure}{l}{25mm}
    \centering
    \includegraphics[width=1in,height=1.25in,clip,keepaspectratio]{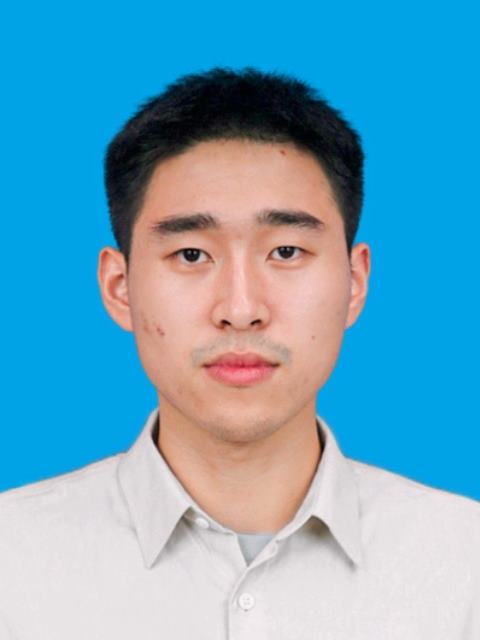}
\end{wrapfigure}
\noindent \textbf{Haoming Chen} is a Ph.D. student in the School of Computer Science and Technology at East China Normal University. His research areas include 2D human pose estimation and 3D large-scale scene perception.\par

\end{document}